%% file: main.tex
\documentclass[preprint,review,12pt]{elsarticle}

\usepackage{amssymb}
\usepackage{amsthm}

\usepackage{comment}
\usepackage{booktabs}
\usepackage{verbatim}
\usepackage{amsmath}
\usepackage{amsfonts}
\usepackage{enumerate}
\usepackage{enumitem}
\usepackage{multirow}
\usepackage{threeparttable}
\usepackage{makecell}
\usepackage{subcaption}
\usepackage{tabularx}
\usepackage{algorithm, algorithmic}
\usepackage{wrapfig}
\usepackage{float}
\newtheorem{theorem}{Theorem}[section]
\newtheorem{corollary}[theorem]{Corollary}

\journal{}

\begin{document}

\begin{frontmatter}



\title{An Efficient Continuous Control Perspective for Reinforcement-Learning-based Sequential Recommendation}


\author[1]{Jun Wang}
\ead{jwang692-c@my.cityu.edu.hk}

\author[2]{Likang Wu}
\ead{wulk@mail.ustc.edu.cn}

\author[3,4]{Qi Liu}
\ead{qiliuql@ustc.edu.cn}

\author[1]{Yu Yang\corref{cor1}}
\ead{yuyang@cityu.edu.hk}

\affiliation[1]{
    organization={School of Data Science, City University of Hong Kong},
    city={Kowloon},
    country={Hong Kong},
}
\affiliation[2]{
    organization={College of Management and Economics, Tianjin University},
    state={Tianjin},
    country={China},
}
\affiliation[3]{
    organization={Anhui Province Key Laboratory of Big Data Analysis and Application, University of Science and Technology of China},
    city={Hefei},
    state={Anhui},
    country={China},
}
\affiliation[4]{
    organization={State Key Laboratory of Cognitive Intelligence},
    city={Hefei},
    state={Anhui},
    country={China},
}
\cortext[cor1]{corresponding author}

\begin{abstract}
Sequential recommendation, where user preference is dynamically inferred from sequential historical behaviors, is a critical task in recommender systems (RSs).
To further optimize long-term user engagement, offline reinforcement-learning-based RSs have become a mainstream technique as they provide an additional advantage in avoiding global explorations that may harm online users' experiences.
However, previous studies mainly focus on discrete action and policy spaces, which might have difficulties in handling dramatically growing items efficiently.

To mitigate this issue, in this paper, we aim to design an algorithmic framework applicable to continuous policies. To facilitate the control in the low-dimensional but dense user preference space, we propose an \underline{\textbf{E}}fficient \underline{\textbf{Co}}ntinuous \underline{\textbf{C}}ontrol framework (ECoC).
Based on a statistically tested assumption, we first propose the novel unified action representation abstracted from normalized user and item spaces.
Then, we develop the corresponding policy evaluation and policy improvement procedures.
During this process, strategic exploration and directional control in terms of unified actions are carefully designed and crucial to final recommendation decisions.
Moreover, beneficial from unified actions, the conservatism regularization for policies and value functions are combined and perfectly compatible with the continuous framework. The resulting dual regularization ensures the successful offline training of RL-based recommendation policies.
Finally, we conduct extensive experiments to validate the effectiveness of our framework.
The results show that compared to the discrete baselines, our ECoC is trained far more efficiently. Meanwhile, the final policies outperform baselines in both capturing the offline data and gaining long-term rewards.
\end{abstract}




\end{frontmatter}


\input{section-I-introduction}

\input{section-II-literature}
\input{section-III-preliminaries}
\input{section-IV-method}
\input{section-V-experiments}

\section{Conclusion}\label{sec_conclusion}

In this paper,  we have studied the problem of RL-based sequential recommendation from the perspective of continuous control, where an efficient framework ECoC is proposed.
Equipped with a brand new action space abstracted from directional user preference and latent item spaces, the policy evaluation and policy improvement procedures of ECoC are redesigned. In particular, the strategic exploration that leverages unified evaluation, as well as deterministic directional policy gradients, plays a key role in enhancing efficiency and reducing model complexity.
Moreover, the dual conservatism regularization is perfectly integrated into ECoC, which leads to a groundbreaking approach in the RL-based recommendation realm.
Last but not least, the comprehensive experiments also confirm that both imitation and off-policy performance of our approach outperform baselines, along with better training efficiency.

In the future, we will explore the stochastic form of the continuous policies in addition to the multi-modal distributions on the high-dimensional latent space, which aims to capture the probabilistic drift of user preference among multiple aspects.

\appendix
\input{section-appendix}



\bibliographystyle{elsarticle-num-names} 
\bibliography{references}

\end{document}

%% file: section-I-introduction.tex
\section{Introduction}

Recommender Systems (RSs) have played an increasingly crucial part in relieving information overload for online services.
Nowadays, user profiles may be unavailable in many scenarios due to privacy limitations and other reasons.
Instead, the historical sequences of user behaviors are accessible and necessary ingredients that can be exploited to infer users' preferences, leading to the sequential recommendation problem.
Under the assumption that users' behaviors exhibit a latent pattern within a certain period, it is feasible to recommend items based on current user interest and further capture pattern transitions across different periods.

Since the Markov property was first introduced into sequential recommendation~\cite{shani2005mdp, rendle2010factorizing}, the dependence of users' next behavior on previous items has been demonstrated to be effective.
Recently, sequential recommendations have enjoyed significant advances with the emerging neural network structures~\cite{hidasi2015session, li2017neural, tang2018personalized, wu2019session, kang2018self}.
However, most previous studies propose supervised-learning-based (SL-based) methods that optimize the immediate user feedback (e.g., click or purchase) but ignore the long-term user engagement. 
To this end, a technical route of reinforcement-learning-based (RL-based) RSs is investigated. The basic idea is to take into account the dynamic interactions between users and the recommendation agent, thus able to guide users using counterfactual inferring rather than prediction.

Unlike SL-based sequential RSs, an inevitable crux of RL-based ones is the following question: What would happen if one RS recommends an item such that the response cannot be inferred from the previous data of the user? One potential solution relies on the global exploration of RL agents, which may harm the experiences of online users.
Thus, researchers turn to offline algorithms and directly study policies from logged data to avoid interacting with online users.
Among these offline methods, both model-based~\cite{shi2019virtual, chen2019generative, zou2020pseudo, chen2021generative, gao2023alleviating} and model-free~\cite{xin2020self, xiao2021general, xin2022supervised, xin2022rethinking} approaches are sought. In this paper, we concentrate on the latter.

Practical recommender systems often encounter an enormous number of items, making the action space very large and the resulting policies with high complexity. The problem becomes even more severe for RL methods due to the widely adopted actor-critic framework, where an additional critic component that takes responsibility for learning the value functions is considered on top of the regular actor component that learns the policy. In particular, the existence of the value function as well as the target update style, leads to the complexity issue being amplified.
Referring to supervised learning methods~\cite{chen2021survey}, an effective solution is to map the large discrete action space into a smaller continuous one.
In this way, manipulating latent representations of user preference becomes the key point, which decides the final concrete items jointly with latent item embedding vectors.
With this motivation, we aim to facilitate the efficient learning of continuous RL-based recommendation policies, which we would highlight to be the first attempt. 


However, managing such a low-dimensional but infinite space is quite challenging. We discover at least the following two obstacles.
First, it is tough to find a well-defined continuous action space.
On the one hand, if we follow the SL-based approaches and define the action space to be composed of user preference representations, the policy evaluation might be problematic.
The reason behind is that the learning of value function depends heavily on strategic exploration in the action space~\cite{fortunato2018noisy, plappert2018parameter}.
When explicit user and item profiles are available, the search procedure is possible through the user-user~\cite{ye2011exploiting} or user-item-user~\cite{jamali2009trustwalker} similarities. However, in sequential scenarios where these profiles are not guaranteed, how to design an effective exploration strategy remains an open question.
On the other hand, if we instead define the action from latent item space, the parameter gradients from the value function might be difficult to be backpropagated to the policy since the existing matching manner relies on the cosine similarity of latent representations.
Second, the issue of extrapolation error caused by limited data coverage in offline scenarios is inevitable~\cite{fujimoto2019off}. Therefore, a penalty of uncertainty either for actions taken by the policy~\cite{xiao2021general, gao2023alleviating} or state-action values~\cite{kumar2020conservative} is compulsory to facilitate the training.
However, it is still doubtful whether these conservatism regularization methods are compatible with the continuous recommendation situation, especially considering the sparsity of user-item interactions.
In essence, we emphasize that all these challenges relate to the interaction mechanism between latent user and item spaces.

To tackle the above challenges, we propose our \underline{\textbf{E}}fficient \underline{\textbf{Co}}ntinuous \underline{\textbf{C}}ontrol framework (ECoC for short), which focuses on accommodating deterministic policies with efficient user preference learning. In summary, our main contributions are as follows.

\begin{itemize}
	\item First, to build the connections between latent user and item spaces in RL-based recommendation, we propose a novel idea of action abstraction to extract unified actions from normalized user and item spaces.
	After statistically testing the significance between ranking orders and angular distances of normalized user/item representations, the reasonability of the abstracted action space is empirically confirmed.


	\item Second, based on the unified action, we develop the concrete procedures of policy evaluation and policy improvement for RL-based sequential recommendation. Specifically, the policy evaluation is realized through strategic exploration within the unified action space, while the policy improvement is also beneficial from the directional update in the same space.
    
    \item Third, the conservatism regularization for both the policy and the value function is perfectly implemented under our ECoC framework and combined to alleviate the extrapolation error.
    In particular, the deterministic policy gradient theorem regarding the unified action is extended to the constrained situation for offline training.

	\item Last but not least, we conduct extensive experiments to evaluate both the imitation and off-policy performance of our proposed framework.
	The results clearly demonstrate that with higher training efficiency, ECoC outperforms the baselines in both mimicking logged data and gaining cumulative long-term rewards under the simulated environment.
\end{itemize}

%% file: section-II-literature.tex
\section{Related Work}
    The two most relevant technical points of this paper are sequential recommender systems (RSs) and offline reinforcement learning. The recent literature on these areas is discussed below.

\subsection{Sequential Recommender Systems}
    Sequential recommendation~\cite{fang2020deep, wang2021survey} is a quite different recommendation scenario from the general case. The basic distinction lies in the absence of explicit user and item portraits. Without such features, the recommendation agent has to capture the dynamic transition patterns in addition to latent representations both for users and items through their interactions. As a result, the basic objective is to recommend the most likely item that appears in the next position within the current session.
    Previous next-item sequential RSs can be mainly categorized into two groups: 1) supervised-learning-based (SL-based) and 2) reinforcement-learning-based (RL-based) methods.
    
    \subsubsection{SL-based Methods}
    At the very beginning, matrix factorization~\cite{rendle2010factorizing} is the most effective approach in conjunction with the Markov assumption. Then, the Recurrent Neural Networks (RNN), as an appealing technique, is beneficial to the construction of more powerful session-based RSs. For example, the most famous solutions include GRU4REC~\cite{hidasi2015session} and its successors, e.g., NARM~\cite{li2017neural}, STAMP~\cite{liu2018stamp} and etc.~\cite{quadrana2017personalizing, li2018learning}. The most obvious advancement lies in the more sophisticated way in which the long-term user interests and short-term preference shifts are combined.
	
    Apart from RNN, another potential structure adopted to recognize temporal patterns is Graph Neural Networks (GNN). The core idea is to regard various session sequences that are composed of items as a huge item graph with transitions as potential edges. Starting from the work of SR-GNN~\cite{wu2019session} and GC-SAN~\cite{xu2019graph}, a bunch of methods~\cite{wang2020global, xia2021self1, xia2021self2} are proposed to further explore either the high-order item-item relations or sequential dependencies in the item graph. 
    
    Moreover, some other structures are investigated to facilitate the session-based recommendation, like the convolutional neural networks (CNNs)~\cite{tang2018personalized, yuan2019simple} and Transformer-based structure~\cite{kang2018self, wu2020sse, sun2019bert4rec, de2021transformers4rec, deng2024sse4rec}.
    Last but not least, the attention mechanism is of particular significance in the sequential recommendation, either for efficient readout designs~\cite{zhang2023efficiently, yuan2021dual} or additional memory modules. The intrinsic motivation is to explicitly mimic the human reasoning process and improve computational efficiency.
    Finally, for the sake of handling popularity bias, the technique of causal inference has presented a strong potential~\cite{gupta2021causer}. 

	However, as concluded in \cite{garg2020batch, xiao2021general}, these supervised methods are actually recovering the behavior policy that generates the logged data. The reason is that only the instantaneous signals are leveraged during training. On the contrary, the way to attain a better policy instead of mimicking the existing one is ignored.
    
  \subsubsection{RL-based Methods}
    As mentioned above, to further learn recommendation strategies that perform better in terms of long-term user engagement, it is promising to incorporate a reinforcement learning framework that aims to interactively model the dynamic process between recommendation agents and users~\cite{chen2021survey}.
	This is exactly the goal of RL-based sequential recommendation agents.
	
    However, due to costly online trials with real users, almost all RL solutions in this scenario seek optimal recommendation policies concerning logged data, which can be further divided into model-based and model-free ones. On the one hand, the former focuses on the construction of the environment, which essentially is a simulated user model, either in an explicit~\cite{chen2019generative, shi2019virtual, zou2020pseudo} or implicit~\cite{chen2021generative, gao2023alleviating} way. As a result, the collection of on-policy transitions and the trial-and-error process can be conducted with this virtual user.
    On the other hand, without permission to recover an environment, which is a time-consuming process, a few researchers~\cite{xin2020self, xiao2021general, xin2022supervised, xin2022rethinking} propose to optimize policies directly by exploring the behavioral ones implied in the data. In such a process, distribution shift is a common challenge since extra on-policy data collection is infeasible.
    Still, other investigations are reported, especially the effects of side information extracted from knowledge graph \cite{wang2020kerl}.
    
    Formally, our work belongs to the model-free methods, where the main difference from the previous ones is the definition of action space and the accompanying policy. Rather than regarding the entire item space as a discrete one, we alternatively take into account the latent spaces which are composed of user preference and item representations, thus making the fine-grained continuous policy learning and transferring possible.

  \subsection{Offline Reinforcement Learning}
    
    Different from classic trial-and-error learning schema in the area of reinforcement learning, offline RL (also known as batch RL)~\cite{levine2020offline, prudencio2023survey} is a paradigm that learns exclusively from static datasets of previously collected trajectories. As such, it can learn policies from diverse experiences, which is quite similar to off-policy RL. However, the prominent obstacle of offline RL is finding a subtle balance between the generalization of potential policies contained in offline data and conservative control of the extrapolation error due to distribution shift~\cite{siegel2020keep}, which is actually caused by the insufficient coverage of offline data~\cite{wang2021statistical}.
	
	In practice, to relieve the distribution shift, a variety of constraints~\cite{wu2019behavior} are adopted during policy improvement and evaluation.
	Specifically, for value-based RL, a pessimistic value function~\cite{kumar2019stabilizing, kumar2020conservative} is proposed to regularize the estimation of the values.
	While for policy-based RL, constraints are imposed through either preventing out-of-distribution actions~\cite{fujimoto2019off, siegel2020keep} or modeling the confidence sets for uncertain regions of state-action pairs~\cite{osband2016deep, o2018uncertainty}.


%% file: section-III-preliminaries.tex
\begin{figure}[ht]
	\centering
	\includegraphics[width=0.95\columnwidth]{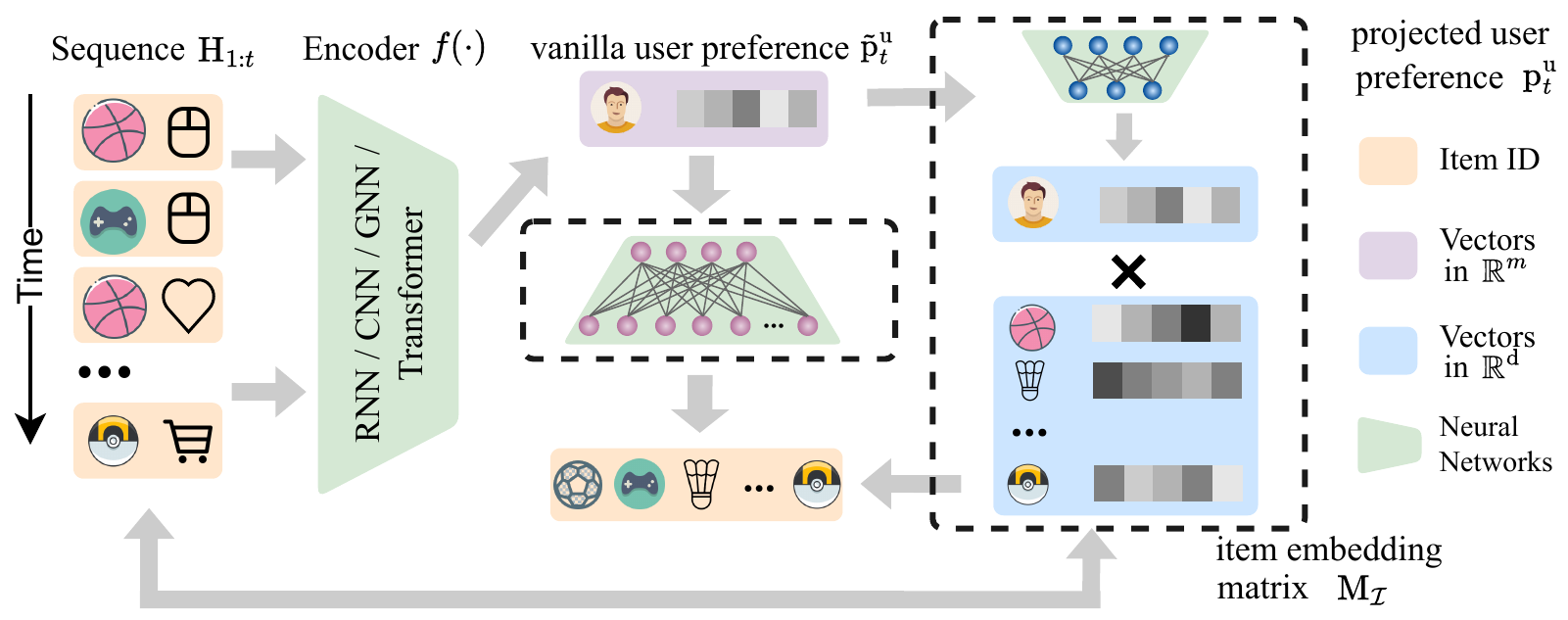}
	\caption{Two implementation manners for sequential recommendation. The discrete version is presented in the dashed box on the left-hand side whereas the continuous version is illustrated in the dashed box on the right-hand side.}
	\label{fig_flow}
\end{figure}

\section{Preliminaries}\label{sec_prelim}

\subsection{Problem Statement}\label{supervised_flow}
	\sloppy This paper aims to solve the next-item sequential recommendation strategy.
    Let $\mathcal{I}$ be the set of all items and $i$ be a certain item, whereas $\mathcal{B}$ denotes the set of all kinds of feedback and $b$ denotes one type. A sequence $\mathrm{H}_{1:t}=\{(i_1, b_i), (i_2, b_2), \cdots, (i_t, b_t)\}$ contains $t$ consecutive behaviors and the sequence buffer $\mathcal{H}=\cap_{k=1}^{K} \mathrm{H}_k$ contains all historical sequences. Given $\mathcal{H}$, our goal is to predict the most rewarding item $i_{t+1}$ at the next position.
    Roughly, there are three steps as illustrated in Figure~\ref{fig_flow}, while the important notations are summarized in Table~\ref{tab_notations}.

    \subsubsection{Embedding Operation: $i \rightarrow \mathrm{e}^i$}
		The embedding matrix $\mathrm{M}_{\mathcal{I}}$ stores the low-dimensional dense representation $\mathrm{e}_i \in \mathbb{R}^{\mathrm{d}}$ of each item $i$, where $d$ is the dimension of latent item space. 

	\subsubsection{Preference Inferring: $\mathrm{H}^{\mathrm{u}}_{1:t} \rightarrow \tilde{\mathrm{p}}^{\mathrm{u}}_{t}$}
        With embedding vectors of items until timestep $t$, supervised learning methods~\cite{hidasi2015session, li2017neural, wu2019session, tang2018personalized, kang2018self, xia2021self2} dwell in constructing an effective encoder $f(\cdot)$ in order to model the user preference as another dense vector $\tilde{\mathrm{p}}^{\mathrm{u}}_{t}$, i.e.,
		\begin{equation*}
			\tilde{\mathrm{p}}_t^{\mathrm{u}} = f( \mathrm{e}^i_1, \mathrm{e}^i_2, \cdots, \mathrm{e}^i_t).
		\end{equation*}
        Generally, we do not expect any dimensional alignment between user preference and item embedding spaces here. Therefore, the vanilla user preference is assumed with dimension $m$.

	\subsubsection{Decision: $ \tilde{\mathrm{p}}^{\mathrm{u}}_{t} \rightarrow i_{t+1}$}
        The decision principle can be mainly classified into either continuous or discrete versions.
		In supervised learning methods, the continuous version is preferred where the major idea is to obtain an effective projection mapping $g(\cdot)$ from $\mathbb{R}^m$ to $\mathbb{R}^{\mathrm{d}}$.
        With this $g(\cdot)$ at hand, $\tilde{\mathrm{p}}^{\mathrm{u}}_t$ is converted into $\mathrm{p}^{\mathrm{u}}_t$ in order to match $\mathrm{e}^i$s.
        Then, decided jointly by $\mathrm{p}^{\mathrm{u}}_t$ and the item embedding matrix $\mathrm{M}_{\mathcal{I}}$, the ranking order is up to the utility scores where the concrete form is often chosen as the inner product~\cite{koren2009matrix}. Consequently, the final top item for recommendation is
		\begin{equation*}
			i_{t+1} 
            = \operatorname{argmax} _{i \in \mathcal{I}} \mathrm{M}_{\mathcal{I}}[i] \cdot \mathrm{p}^{\mathrm{u}}_t,
		\end{equation*}
        where $\mathrm{M}_{\mathcal{I}}[i]$ denotes the $i$-th row of $\mathrm{M}_{\mathcal{I}}$, which is actually $\mathrm{e}^i$.
        In practice, we typically select the top-$N$ items with the highest utilities.

        Alternatively, for those RL-based models~\cite{xin2020self, xiao2021general, xin2022supervised}, to coordinate with the value function (as detailed in Section~\ref{sec_overview}), the discrete version is more widely-adopted where the decision can be directly expressed as $g(\cdot): \mathbb{R}^m \mapsto \mathbb{R}^{|\mathcal{I}|}$, where $g(\cdot)[i]$ represents a score, probability or ranking order of item $i$ to occur at the next position and $$i_{t+1} =  \operatorname{argmax} _{i \in \mathcal{I}} g(\tilde{\mathrm{p}}^{\mathrm{u}}_t)[i].$$

	\subsubsection{Complexity Comparison}\label{sec_actor_complex}
        Obviously, compared to the discrete framework, the continuous version in supervised learning methods is more attractive since it exhibits a significant superiority w.r.t. the model complexity.
        Taking the simplest one-layer fully-connected neural network as an example, the model complexity of the approximator of $g(\cdot)$ is reduced from $\mathcal{O}(m |\mathcal{I}|)$ to $\mathcal{O}(md)$ where $d << |\mathcal{I}|$ in practice.
        Therefore, our work aims to keep this complexity superiority when designing RL-based recommendation frameworks.

\begin{table}
	\centering\footnotesize
	\caption{Important Mathematical Notations.}
	\begin{tabular}{>{\centering}p{0.04\textwidth}|p{0.08\textwidth}p{0.85\textwidth}}
		\toprule
		\multicolumn{1}{c}{} &  Symbol & \multicolumn{1}{c}{Description} \\
		\midrule
		\multirow{13}{*}{\rotatebox[origin=c]{90}{Basic}} & $\mathcal{I}$  & set of all items, with each represented by an index $i$ \\
		& $\mathrm{H}$  & a behavior sequence containing historical items and (optional) feedback \\
		& $\mathcal{H}$  & offline dataset containing all behavior sequences, i.e.,  $\mathcal{H}=\cap_{k=1}^{K} \mathrm{H}_k$\\
		& $\mathrm{e}^i (\bar{\mathrm{e}}^i)$  & the (normalized) embedding vector of item $i \in \mathcal{I}$, with $||\bar{\mathrm{e}}^i||_2=1$ \\
		& $\mathrm{M}_{\mathcal{I}}$ & the item embedding matrix, including all $\mathrm{e}^i$s, of which the shape is $|\mathcal{I}| \times d$ \\ 
		& $\overline{\mathrm{M}}_{\mathcal{I}}$ & normalized item embedding matrix, including all $\bar{\mathrm{e}}^i$s \\
		& $d$  & the dimension of latent item space \\
		& $m$  & the dimension of vanilla user preference space \\
		& $\tilde{\mathrm{p}}^{\mathrm{u}}$  &  vanilla user preference vectors in $ \mathbb{R}^m$ \\
		& $\mathrm{p}^{\mathrm{u}} (\bar{\mathrm{p}}^{\mathrm{u}})$  &  (normalized) projected user preference vectors in $ \mathbb{R}^{\mathrm{d}}$, with $||\bar{\mathrm{p}}^{\mathrm{u}}||_2=1$ \\
		& $f(\cdot)$	& the encoder that extracts vanilla user preference $\tilde{\mathrm{p}}^{\mathrm{u}}$ from the sequence $\mathcal{H}$	\\
		& $g(\cdot)$	& the decision function, of which the concrete form depends on either discrete or continuous manner	\\
		\midrule
		\multirow{5}{*}{\rotatebox[origin=c]{90}{test-related}} & $X$ & the vanilla inner product between $\mathrm{M}_{\mathcal{I}}$ and $\mathrm{p}^{\mathrm{u}}$ \\
		& $Y$ & the inner product between $\overline{\mathrm{M}}_{\mathcal{I}}$ and $\bar{\mathrm{p}}^{\mathrm{u}}$ \\
		& $Z$ & the norm product between $\mathrm{M}_{\mathcal{I}}$ and $\mathrm{p}^{\mathrm{u}}$ \\
		& $\rho^{XY}$ & the Spearman correlation coefficient between $X$ and $Y$ \\
		& $z^{XY}$ & the outcome of Fisher z-transformation on $\rho^{XY}$ \\
		\midrule
		\multirow{6}{*}{\rotatebox[origin=c]{90}{RL-related}} & $s_t$  &  the state at timestep $t$, actually the same as $\mathrm{H}_{1:t}$ \\
		& $a$   & the action abstracted from either $\bar{\mathrm{e}}^i $ or $\bar{\mathrm{p}}^{\mathrm{u}}$, with $||a||_2=1$ \\
		& $\mu(\cdot) $	& the user preference extraction function, which outputs $\mathrm{p}^{\mathrm{u}}$	\\
        & $\pi(\cdot) $	& the policy extending from $\mu(\cdot)$, which outputs $a$	\\
		& $Q(s, a)$  & the value decided by the critic when the actor takes action $a$ for state $s$ \\
		& $N_1$ & the sampling parameter in $\mathcal{L}_{REG}$ \\
		& $N_2$ & the sampling parameter in $\mathcal{L}_{BC}$ \\
		\bottomrule
	\end{tabular}%
	\label{tab_notations}%
\end{table}%

\subsection{MDP Formulation}\label{rl_formula}
    Before we introduce our practical actor-critic-based framework, we first formulate the interactions of recommendation agents and users as a Markov Decision Process (MDP), i.e., a quadruple $\mathcal{M} =(\mathcal{S}, \mathcal{A}, \mathcal{P}, \mathcal{R})$~\footnote{Note that here the symbol $\mathcal{M}$ is distinguished with the item embedding matrix $\mathrm{M}_{\mathcal{I}}$.}.

\begin{itemize}
	\item \textit{Environment}. The environment corresponds to a user model that provides different feedback to the displayed items, like clicking or purchasing, in accordance with the logged data.
	
	\item \textit{State: $\mathcal{S}$}. Under the assumption that a user's latent preference can be perceived through the observable historical behaviors, we consider a sequence $\mathrm{H}_{1:t}$ as the state $s_t$ at time $t$.
	
	\item \textit{Action: $\mathcal{A}$}.
    Ultimately, the final recommendation outcomes are discrete indices of items. However, since we aim to manage fine-grained control of latent representations, the concrete meaning of actions, as well as the policy, will be detailed in Section~\ref{sec_action}. 
    Absolutely, it is a low-dimensional but infinite space.
	
	\item \textit{State Transition: $\mathcal{P}$}.
	The transition probability function corresponds to a subsequent user behavior sequence $\mathrm{H}_{1:t+1}$, which covers $\mathrm{H}_{1:t}$ as well as the following pair of item-feedback $(i_{t+1}, b_{t+1})$.
	
	\item \textit{Reward Function: $\mathcal{R}$}.
	The reward signals are exactly the users' feedback on the displayed items, which depend on the logged datasets. We explain the practical setting in Section~\ref{sec_eval_metric}. Basically, the more positive the feedback is, the higher the value returned by the function. However, in most cases, rewarding feedback is infrequent, which makes the extrapolation error problem more severe when evaluating policies.
\end{itemize}
	
	
	Finally, the objective of RL-based recommendation is to learn an optimal policy (i.e., recommendation strategy) that maximizes the expected cumulative rewards
	$
		\mathbb{E} [ \sum \nolimits_{t=0}^N \gamma ^t r(s_t, a_t)].
	$
    When the discount factor $\gamma \rightarrow 0$, the policy is asymptotically equivalent to the one trained by SL methods.

\section{Representation Normalization: Assumption and Statistical Test}\label{sec_aasumption}
This section will introduce an essential step in our proposed method, i.e., representation normalization.
Our idea is to share only the angular information across the latent item and user preference spaces and obtain a unified description abstracted from these two spaces, which is crucial to both policy evaluation and policy improvement. We will first illustrate the intuitive motivation for normalizing representations, and then develop a paired sample t-test to verify the significance of this operation. 
    
\subsection{Intuitive Motivation for Normalizing Representations}
    
As increasingly adopted in previous methods~\cite{abdollahpouri2017controlling, gupta2019niser, zhang2023efficiently}, the normalization of item representations is an effective operation to control the popularity bias that is common in recommendation scenarios. Such a viewpoint of decomposing the inner product into $L_2$ norm and angular distance raises a further issue of to what degree the angular distance contributes to the final ranking scores.
    
    In addition, except for the limited application of normalization in GNN-based models~\cite{gupta2019niser, zhang2023efficiently}, we hope to perform a broader validation of whether the ranking order is fairly preserved by the inner products (i.e., cosine similarities) of normalized user and item representations.
    The reason why we focus on this question is to explore a unified description technique across such two spaces. While the distribution of $L_2$ norm values might vary depending on the intrinsic property of latent spaces, it is more hopeful that the directional representations with norm standardized imply inside allocation proportions and thus can be shared.
	
	To figure out the attribution of inner-product-based ranking strategies, we formally claim the hypothesis below and design a paired Student's t-test in the following subsection.
    The null hypothesis is formulated as
    \begin{gather*}
        \mathbb{H}_0: \textit{\text{the angular distance of $\mathrm{e}^i$ and $\mathrm{p}^{\mathrm{u}}$ is less important than}}\\
        \textit{\text{the product of $L_2$ norm in terms of ranking by $\mathrm{M}_{\mathcal{I}} \cdot \mathrm{p}^{\mathrm{u}}$}},
    \end{gather*}
    against the alternative hypothesis being
    \begin{gather*}
        \mathbb{H}_1: \textit{\text{the angular distance of $\mathrm{e}^i$ and $\mathrm{p}^{\mathrm{u}}$ is more important than}} \\
        \textit{\text{the product of $L_2$ norm in terms of ranking by $\mathrm{M}_{\mathcal{I}} \cdot \mathrm{p}^{\mathrm{u}}$}}.
    \end{gather*}
    Here, we investigate the importance of such two factors by studying the correlation with the final ranking orders.

    \subsection{Statistical Test}
    
    To obtain the final test samples, we first state the following notations
    \begin{equation*}
        \begin{cases}
        \mathrm{e}^i = \bar{\mathrm{e}}^i \cdot \omega ^i, \quad \text{with}  \quad \omega ^i = ||\mathrm{e}^i||_2, \\
        \mathrm{p}^{\mathrm{u}} = \bar{\mathrm{p}}^{\mathrm{u}} \cdot \omega ^{\mathrm{u}}, \quad  \text{with}  \quad \omega ^{\mathrm{u}} = ||\mathrm{p}^{\mathrm{u}}||_2,
        \end{cases}
    \end{equation*}
    where the norm and direction of user (item) vectors are explicitly represented. In addition, $\overline{\mathrm{M}} _{\mathcal{I}}$ is the normalized item matrix composed of all $\bar{\mathrm{e}}^i$s.
    
    Then, we compute \textbf{Spearman's rank correlations} among the vanilla inner product $X$, the normalized inner product $Y$, and the signed norm vector of all items $Z$ for a given user preference $\mathrm{p}^{\mathrm{u}}$. We hereby moderately assume that all $\mathrm{p}^{\mathrm{u}}$ vectors are generated independently and identically from a $d$-dimensional multivariate normal distribution.
    After defining $\omega_{\mathcal{I}}=(\omega ^i)_{i \in \mathcal{I}}$, we have
    \begin{equation*}
        X= \mathrm{M}_{\mathcal{I}} \cdot \mathrm{p}^{\mathrm{u}},
        \quad Y= \overline{\mathrm{M}}_{\mathcal{I}} \cdot \bar{\mathrm{p}}^{\mathrm{u}},
        \quad Z= \omega_{\mathcal{I}} \cdot \omega ^{\mathrm{u}} \otimes \operatorname{sgn}(X),
    \end{equation*}
    \begin{equation*}
        \rho^{XY} = \rho(\text{Rank}(X), \text{Rank}(Y)),
        \quad \rho^{XZ} = \rho(\text{Rank}(X), \text{Rank}(Z)),
    \end{equation*}
    where $\operatorname{sgn}(\cdot)$ is the signum function, $\otimes$ is Hadamard product, and $\text{Rank}(\cdot)$ denotes the ranking order of its internal elements given a vector.
    With the help of Fisher z-transformation~\cite{fisher1915frequency}, i.e., $z=\frac{1}{2} \ln \left( \frac{1+\rho}{1-\rho} \right)$, the correlation coefficients can be further converted into variables approximately normally distributed. Then, we conduct the paired Student's t-test between $z^{XY}$ and $z^{XZ}$ to test whether their means are statistically equal.

    In experiments, we perform the test on a wide range of supervised methods, including RNN-based (GRU4Rec, NARM, STAMP), CNN-based (Caser, NextItNet), GNN-based (SR-GNN, GC-SAN, DHCN) and Transformer-based (SASRec, SSE-PT, BERT4Rec).
    For each method, we first finish the continuous end-to-end training process detailed in Section~\ref{supervised_flow}.
    Next, we randomly sample $N$=10000 interactions, each with a historical sequence and the ground-truth next item from test data. We then utilize the well-trained models to acquire the differences between all $z^{XY}$ and $z^{XZ}$, along with the mean and standard deviation.
    Such a process is repeated 20 times to obtain the average statistical values.
    The preparation of test data, as well as some implementation details, can refer to Section~\ref{sec_exp_setup}.

    

\begin{table}[ht]
	\centering\footnotesize
	\caption{Testing Results of Paired Sample t-Test between $z^{XY}$ and $z^{XZ}$ with N=1000}\label{tab_embed}
	\setlength{\tabcolsep}{5.pt}
	\begin{threeparttable}
		\begin{tabular}{l||ccc|ccc|ccc}
			\toprule
			\multicolumn{1}{c}{\multirow{2}{*}{Method}}  & \multicolumn{3}{c}{Yelp} & \multicolumn{3}{c}{Tmall}  &  \multicolumn{3}{c}{RL4RS-A}  \\
			\cmidrule(lr){2-4} \cmidrule(lr){5-7} \cmidrule(lr){8-10} \multicolumn{1}{c}{} & Mean & Std. & t & Mean & Std. & t & Mean & Std. & t \\
			\midrule
			GRU4Rec & 2.26 & 0.14 & 1624* & 2.25 & 0.07 & 3029* & 2.08 & 0.06 & 3211* \\
			NARM  & 2.23 & 0.13 & 1669* & 2.22 & 0.06 & 3478* & 2.10 & 0.07 & 3128* \\
			STAMP & 2.21 & 0.12 & 1741* & 2.24 & 0.06 & 3618* & 2.07 & 0.06 & 3348* \\
			\midrule
			Caser & 1.90 & 0.18 & 1045* & 1.45 & 0.10 & 1385*  & 1.33 & 0.13 & 1078* \\
			NextItNet & 1.86 & 0.19  & 989* & 1.32 & 0.12 & 1066*  & 1.28 & 0.12 & 996* \\
			\midrule
			SR-GNN & 2.45 & 0.13 & 1891* & 2.33 & 0.07 & 3298*  & 2.11 & 0.07 & 3114*\\
			GC-SAN & 2.42 & 0.12 & 1998* & 2.31 & 0.06 & 3750* & 2.15 & 0.08 & 2862* \\
			DHCN & 2.43 & 0.14 & 1835*  & 2.34 & 0.07 & 3442*  & 2.14 & 0.07 & 3057* \\
			\midrule
			SASRec & 2.48 & 0.12 & 2098* & 2.30 & 0.05 & 4595* & 2.12 & 0.07 & 3063* \\
			SSE-PT & 2.49 & 0.13 & 1924* & 2.32 & 0.06 & 3966* & 2.13 & 0.07 & 3102* \\
			BERT4Rec & 2.47 & 0.12 & 2035* & 2.29 & 0.05 & 4317* & 2.12 & 0.08 & 2850* \\
			\bottomrule
		\end{tabular}%
		\begin{tablenotes}
			\item * means the P-value is $<$ 0.001, which is highly significant.
		\end{tablenotes}
	\end{threeparttable}
\end{table}%

\begin{figure}[ht]
	\centering
	\includegraphics[width=.65\columnwidth]{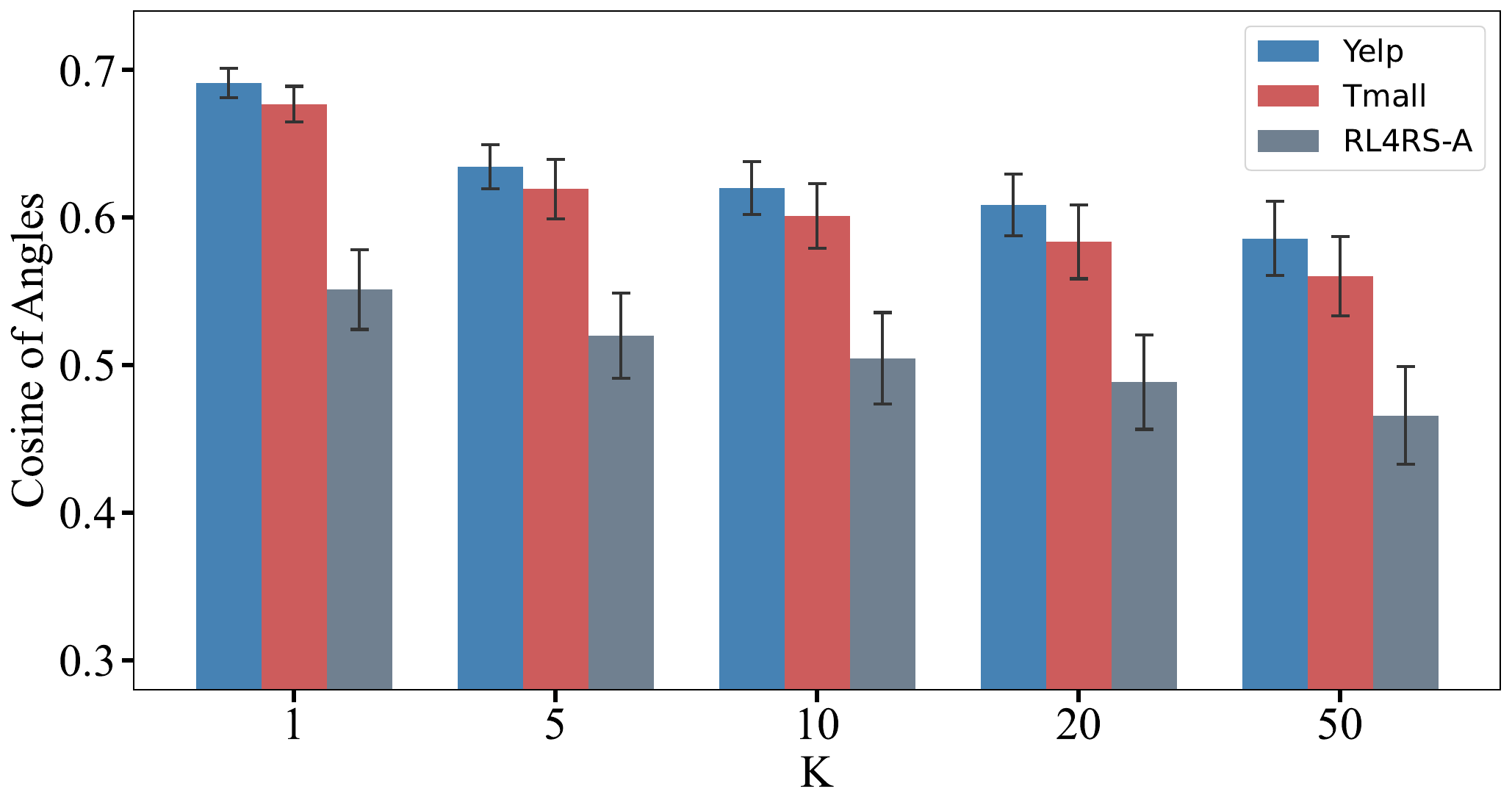}
	\caption{Cosine values of angles between $\mathrm{e}^i$ and $\mathrm{p}^{\mathrm{u}}$ when the ground-truth item $i$ ranks in top-K.}
	\label{fig_cosine}
\end{figure}

    \noindent \textbf{Testing Results.}
    The statistical results are summarized in Table~\ref{tab_embed}. Although the structures utilized are various, we find that our null hypothesis can always be significantly rejected.
    As a result, the alternative hypothesis is statistically accepted, which means the angular distance of normalized embedding vectors across user and item spaces actually preserves adequate evidence regarding the ranking orders.

    In addition, to empirically validate the effectiveness of the proposed unified evaluation technique, we also report the cosine values of angles between latent user preference vectors and ground-truth item embedding vectors regarding the same samples for statistical testing.
    As shown in Figure~\ref{fig_cosine}, there appears to be a trend that the higher ranking order ground-truth items on the test data rank among all candidates, the smaller the angles between $\mathrm{e}^i$ and $\mathrm{p}^{\mathrm{u}}$ are, which means such two latent spaces share a fairly consistent tendency.
    Therefore, it is possible to abstract unified representations from normalized $\mathrm{e}^i$ or $\mathrm{p}^{\mathrm{u}}$ vectors.

%% file: section-IV-method.tex
\begin{figure}[ht]
	\centering
	\includegraphics[width=1.\columnwidth]{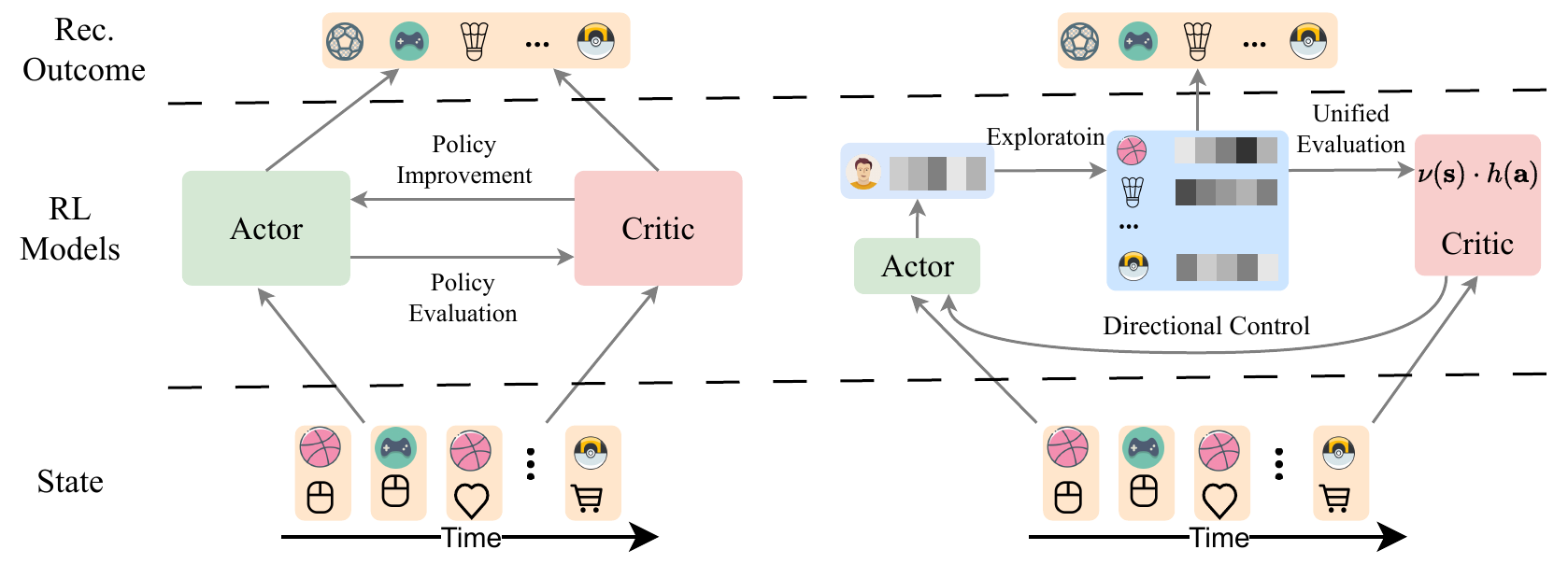}
	\caption{The existing discrete version of actor-critic recommendation framework (\textbf{Left}) and our ECoC framework (\textbf{Right}), which is designed for continuous action spaces and policies. The key difference lies in the utilization of the item embedding matrix, which leads to the unified evaluation and directional control.}
	\label{fig_ecoc}
\end{figure}

\section{The Proposed Framework}
    After introducing the high-level structure of our framework, we focus on the details of the abstracted action space, followed by the procedure of policy evaluation and policy improvement. 
    Moreover, the conservatism regularization is also perfectly integrated to relieve the extrapolation error.
	
\subsection{Framework Overview}~\label{sec_overview}
    As illustrated in Figure~\ref{fig_ecoc}, this paper adopts the actor-critic framework at a high level, where the actor learns a policy $\pi(\cdot)$ (parameterized by $\theta$) mapping from the state space to the action space and the critic is responsible for judging the state-action $(s, a)$ pairs through a $Q$ value function (parameterized by $\phi$).
    Basically, the optimization process can be split into two iterative procedures, i.e., policy evaluation and policy improvement.
    Specifically, the policy evaluation refers to the self-learning of the critic, which is exactly the recursive iterations from the Bellman equation~\cite{sutton2018reinforcement} and can be described as
    \begin{equation}\label{eq_critic_generic}
		Q^{\pi}\left(s_t,a_t \right) = \mathbb{E} _{s_{t+1} \sim \mathcal{P}, a_{t+1} \sim \pi} \left[ r + \gamma Q^{\pi} \left(s_{t+1}, a_{t+1} \right)  \right].
	\end{equation}
    
    With a $Q(s, a)$ function at hand, the continuous version of policy improvement can then be realized by $\nabla_{a} Q(s, a) \mid _{a=\pi(s)}$, where the critic implies the directions that can optimize the $Q$ values.
	
\subsection{Abstracted Action Space}\label{sec_action}
    
    In order to be compatible with the continuous recommendation manner in Section~\ref{supervised_flow}, the action space must be composed of continuous vectors.
    However, if such a space is set as the image of the function
    \begin{equation}
        \mu(s_t) := g \circ f \circ \text{Embedding}(\mathrm{H}_{1:t}),
    \end{equation}
    which includes all user preference $\mathrm{p}^{\mathrm{u}}$s in $\mathbb{R}^{\mathrm{d}}$, it is difficult to find an appropriate $a_{t+1}$ in Equation~\eqref{eq_critic_generic}.
    In other words, we need to find the indices of possible items and construct new sequences, followed by generating user preference from scratch.
    Alternatively, the adoption of the latent item space composed of all $\mathrm{e}^i$s brings trouble in backpropagating gradients $\nabla_{a} Q(s, a) \mid _{a=\pi(s)}$. Currently, taking action in latent item space relies on the greedy search of ranking orders, i.e., $\max_{i \in \mathcal{I}} \mathrm{e}^i \cdot \mathrm{p}^{\mathrm{u}}$, where the gradient propagation in terms of the parameters $\theta$ of $\pi(\cdot)$ is hindered.
    
    Therefore, we aim to seek a unified representation manner and abstract an action space from both user preference and latent item spaces.
    The intuition is to sufficiently leverage the item embedding matrix $\mathrm{M}_{\mathcal{I}}$ which contains rich information, as emphasized in Figure~\ref{fig_ecoc}.
    Yet, we emphasize that the interaction mechanism between the actor and the critic should be redesigned.
    
    Formally, based on the statistical assumption in Section~\ref{sec_aasumption}, we denote the action $a$ as a normalized continuous vector from either item embedding or user preference space, i.e., $a \in \left\{\bar{\mathrm{p}}^{\mathrm{u}}, \bar{\mathrm{e}}^i \right\}$, where $||a||_2=1$.
    As such, we can slightly extend the original user preference extraction module by defining
    \begin{equation}
        \pi(s_t) = \frac{\mu(s_t)}{||\mu(s_t)||_2}.
    \end{equation}
    Meanwhile, the generation of $a_{t+1}$ in Equation~\eqref{eq_critic_generic} is also facilitated as described in the section below.
    

\subsection{Policy Evaluation}
    \subsubsection{Strategic Exploration}~\label{sec_ee_tradeoff}
	From the generic form of policy evaluation in Equation~\ref{eq_critic_generic}, the generation of next actions $a_{t+1}$ is often the key to facilitate a well-performed critic,
    where the expectation on the state transition $P(s_{t+1} \mid s_t,a_t)$ is often estimated by sampling under the distribution implied in logged data.
    The main advantage of RL-based recommendation policies lies in better long-term optimality under the guidance of reward signals~\cite{xiao2021general}.
    Even for the offline scenario, moderate explorations are critical to the ability of generalization beyond the behavior policy recovered from logged data.
    
    For a continuous action space as well as the deterministic policies, noise-based exploration~\cite{fortunato2018noisy, plappert2018parameter} is a common solution. However, the extremely sparse interactions between users and items hinder the application of such methods. Without any prior knowledge of the distribution of latent user preference space, it is infeasible to generate stochastic noises.
    
    Fortunately, based on the policies and actions defined above, we propose a novel \textbf{unified evaluation} technique to realize the strategic exploration across latent user and item spaces.
    Specifically, the vanilla manner of policy evaluation is altered to criticize abstracted actions.
    Then, due to the close relations between the angular distances of $\mathrm{e}_i$ and $\mathrm{p}^{\mathrm{u}}$ and the ranking orders, the normalized item embedding vectors $\bar{\mathrm{e}}^i$ should store adequate angular information on nearby user preference vectors $\bar{\mathrm{p}}^{\mathrm{u}}$. Therefore, we can leverage these exploration directions indicated by representations of similar items.
    In other words, we impose the exploration of $a_{t+1}$ by perturbing it to be $\bar{\mathrm{e}}^i$ in the neighborhood of $\bar{\mathrm{p}}^{\mathrm{u}}_{t+1}$, where $\mathrm{p}^{\mathrm{u}}_{t+1} = \mu(s_{t+1})$ is the user preference at timestep $t+1$.
    
\begin{figure}[ht]
    \centering
    \begin{minipage}[t]{0.8\columnwidth}
        \centering
        \includegraphics[width=0.42\textwidth]{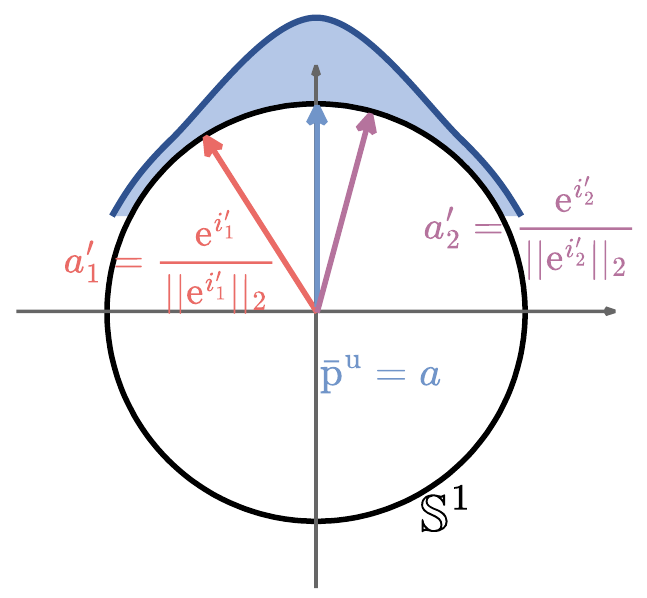}
        \hfill
        \includegraphics[width=0.425\textwidth]{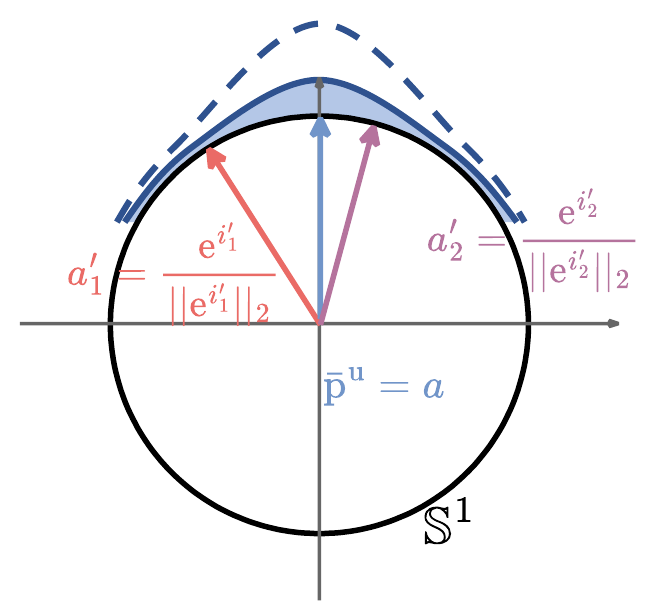}
        \caption{An illustration of \textbf{unified evaluation} (L) and  \textbf{conservatism regularization} (R) when $d=2$.
        }
        \label{fig_unified_eval}
    \end{minipage}
\end{figure}
    
    We provide another \textbf{mathematical illustration} of our unified evaluation technique in Figure~\ref{fig_unified_eval}. Given a von Mises-Fisher distribution~\cite{fisher1953dispersion} on the $(d-1)$-sphere in $\mathbb{R}^{\mathrm{d}}$ of which the probability density is
    $
        f_d (\boldsymbol{x}; \boldsymbol{\nu}, \kappa) = C_d(\kappa) \exp (\kappa \boldsymbol{\nu}^{T} \boldsymbol{x})
    $,
    where $\boldsymbol{\nu}$ is the mean direction, $\kappa$ is the concentration parameter and $C_d(\kappa)$ is a normalization constant. 
    If $\boldsymbol{\nu}$ is supposed as the normalized user preference $\bar{\mathrm{p}}^{\mathrm{u}}$, the latent item space essentially derives a non-parametric estimation of $f_d (\boldsymbol{x}; \bar{\mathrm{p}} ^{\mathrm{u}}, \kappa)$ and thus avoiding estimating the state-dependent parameter $\kappa$ when the number of items is dense enough.

    In practice, incorporating the unified evaluation, the update of the critic is presented as follows.
    Given transitions $B=\left\{ (s_j, i_j, r_j, s_j') \right\}_{j=1}^{|B|}$ sampled from $\mathcal{H}$, the minimization of temporal error from the two sides of Eq.~\eqref{eq_critic_generic} is performed by
    \begin{equation}\label{eq_loss_td}
    \begin{gathered}
        \mathcal{L}_{TD} = \mathbb{E}_{ \left( s, i, r, s' \right) \sim \mathcal{H}} \left[ Q(s, a) - \left( r + \gamma Q(s', a') \right) \right] ^2, \\
        \text{where} \quad a = \frac{\mathrm{e}^i}{||\mathrm{e}^i||_2}, \quad a' = \frac{\mathrm{e}^{i'}}{||\mathrm{e}^{i'}||_2}, \quad i' \sim \xi \left( \mathrm{M}_{\mathcal{I}}, \pi(s') \right).
    \end{gathered}
    \end{equation}
    For the discrete item IDs $i_k$ recorded in logged data, we convert it into the unified action $\mathrm{e}^{i_k} / ||\mathrm{e}^{i_k}||_2$.
    While for exploration when computing $Q(s', a')$, on-policy actions $\pi(s')$ are disturbed to the latent representations of $i$ jointly decided with $\mathrm{M}_{\mathcal{I}}$.
    
    It should be noted that such item IDs are generated by distributional sampling or neighborhood search procedures depending on the efficiency requirement, which is the exact meaning of $\xi(\cdot, \cdot)$.
    To further fend off the overestimation in the Q function, the technique of clipped double-Q learning~\cite{fujimoto2018addressing} is also inherited.
    

    \subsubsection{Conservatism Regularization}\label{sec_critic_conserv}
	
    The inevitable challenge in yielding an excellent critic is the \textbf{extrapolation error}.
	Due to the limited offline data coverage and out-of-distribution (OOD) issue in training, it is impossible to generalize over the complete $(\mathcal{S}, \mathcal{A})$ space, let alone both of them are continuous ones.

    A direct insight is to constrain the output of optimized policies, which is exactly what we do when performing policy improvement below.
    However, we list two reasons why conservatism regularization is also crucial for the critic.
    First, even with a behavioral penalization in improving the actor, the severe sparsity of user-item interactions in logged data, in addition to the end-to-end training paradigm, makes the policy $\pi$ hard to perfectly recover the behavior policy contained in logged data, especially at the very beginning of the training process.
    Second, according to \cite{silver2014deterministic}, the overestimation of $Q$ values still exists, which is caused by the implicit greedy approximation of $Q(s_{t+1}, a_{t+1})$, i.e., $\max_{a_{t+1}}Q(s_{t+1}, a_{t+1}) \approx Q(s_{t+1}, \pi(s_{t+1}))$.
   
    Therefore, our basic intention is to minimize the expectation of Q values under the true policy while particularly maximizing the truly behavioral one at the same time. Thus, we propose to minimize the following term
    \begin{equation}\label{eq_reg_vanilla}
    	\begin{aligned}
    		\mathcal{L}_{REG} &= \mathbb{E}_{a' \sim \pi(s)} Q (s, a') - Q(s, a) \\
    		&=Q (s, \pi(s)) - Q(s, a) \\
    		& \approx Q (s, \pi_{k-1}(s)) - Q(s, a),
    	\end{aligned}
	\end{equation}
	where $a = \frac{\mathrm{e}^i}{||\mathrm{e}^i||_2}$ is the recorded action from logged item $i$.
	Here, the second equation holds when the policy $\pi(\cdot)$ is deterministic, whereas its true form is approximated by the current policy iterate $\pi_{k-1}(\cdot)$.
    
    
    To make the value penalization more stable, we further relax the value of $Q (s, \pi_{k-1}(s))$ in the sense of expectation over the non-parametric distribution illustrated in Figure~\ref{fig_unified_eval}.
    Beneficial from the unified evaluation technique again, we are able to estimate such an expectation by generating $N_1$ abstracted actions around $\pi_{k-1}(s)$, with the help of latent item space. 
    Finally, such regularization loss is formed by
    \begin{equation}\label{eq_final_reg}
    	\begin{gathered}
        \mathcal{L}_{REG} = \mathbb{E} _{\left(s, i\right) \in \sim \mathcal{H}} \left[ \sum_{k=1}^{N_1} \left[Q(s, a'_k) \frac{\exp (Q(s, a'_k))}{Z}\right] - Q(s, a) \right], \\
        \text{where} \quad a = \frac{\mathrm{e}^i}{||\mathrm{e}^i||_2}, \quad a'_k = \frac{\mathrm{e}^{i'_k}}{||\mathrm{e}^{i'_k}||_2}, \quad i'_k \sim \xi \left( \mathrm{M}_{\mathcal{I}}, \pi(s) \right).
    \end{gathered}\end{equation}
    Here, $N_1$ is the sample number and $Z$ is the normalization term $ \sum_{k=1}^{N_1} \exp (Q(s, a'_k))$. Moreover, the second term $Q(s, a)$ encourages the value of ground-truth actions, which just represent the real user preference.

    We notice that this contrastive formulation in $\mathcal{L}_{REG}$ is similar to the idea induced by the traditional pairwise loss, e.g., Bayesian pairwise ranking (BPR) loss~\cite{rendle2009bpr}.
    In essence, they are divergent. While BPR-like ideas aim to maximize the utility gap of positive and generated negative items, some previous RL-based recommender systems~\cite{zhao2018recommendations, xin2022supervised} directly adopted such an inspiration during the training of the critic.
    However, the partial order assumption of user preference on non-interact items, which forms the theoretical basis of BPR, might not be applicable to the value function anymore.
    Since $Q$ estimates the long-term utility, some certain items with implicit feedback might meet better the user preference in long-term performance, i.e., $\exists k, Q(s, a'_k) \geq Q(s, a)$.
    Consequently, our regularization offers a safer alternative to upper-bound those $Q(s, a'_k)$ by their true values, as proved in \cite{kumar2020conservative}.
    Our ablation study in Section~\ref{sec_ablation} also confirms this opinion.
    
 
\subsection{Policy Improvement}

    \subsubsection{Behavioral Constraint}

    Ahead of regular policy guidance from the critic, we have to recover the behavior policy implied in logged data as accurately as possible.
    First, shrinking the action space helps to mitigate the distribution mismatches when estimating the expectation in Eq.~\eqref{eq_loss_td}.
    Secondly, such a constraint term is critical to mildly guarantee the constrained policy space requirement in the following constrained directional control corollary.
    Thirdly, we emphasize that following the idea of self-supervised learning~\cite{xin2020self}, this term facilitates the latent representation learning of item space in the end-to-end continuous framework~\cite{liu2020end}.
    
	 
	Basically, the actor is required to generate continuous actions that result in the maximal likelihood of concrete items in logged data.
    To be specific, the pairwise ranking loss, e.g., BPR loss~\cite{rendle2009bpr}, is taken into account w.r.t. the final ranking scores
    \begin{equation}\label{eq_pairwise}
        \mathcal{L}_{BC} = \mathbb{E} _{(s, i) \sim \mathcal{H}} \frac{1}{N_2} \sum_{\substack{j_k \in \mathcal{I}, \\ j_k \neq i, k=1, \ldots, N_2}} - \ln \sigma \left(y_i - y_{j_k} \right),
    \end{equation}
    where $\{j_k\} _{k=1} ^{N_2}$ are $N_2$ sampled items with negative feedback in terms of state $s$, $y_i\ (y_{j_k})$ is the ranking score of item $i\ (j_k)$ and $\sigma(\cdot)$ is the Sigmoid function.
	 

    \subsubsection{Constrained Directional Control}

    Since we can manage unified evaluation in abstracted action space, it is feasible to leverage the gradient information in latent item space to instruct the update direction of estimated user preference vectors $\nabla_{a} Q(s, a) \mid _{a=\pi(s)=\frac{\mu(s)}{||\mu(s)||_2}}$, which is the first attempt to introduce such a directional control manner.
    
    The remaining obstacle is whether such directions are unbiased due to the existence of limited experiences and our dual conservatism regularization term.
    Fortunately, such a deterministic update principle still holds, which is exactly the corollary below.
    \begin{corollary}\label{corollary_5.1}
        Under an empirical MDP $\widehat{\mathcal{M}}$ derived from logged data $\mathcal{H}$, with deterministic policy $\pi_{\theta}$ (parameterized by $\theta$) from the constrained policy space $\left\{ \pi \mid \left(s, \pi(a) \right) \in \mathcal{H} \right\}$ and the accompanying value function $Q^{\pi}$, if $\nabla _{\theta} \pi _{\theta}(s)$ and $\nabla_{a} Q^{\pi}(s,a)$ exist,
        then the gradient of empirical performance objective $J (\pi_{\theta}, \widehat{\mathcal{M}})$ is
        \begin{equation}
            \nabla _{\theta} J (\pi_{\theta},  \widehat{\mathcal{M}}) = \mathbb{E}_{s \sim \rho^{\pi}} \left[\nabla _{\theta} \pi _{\theta}(s) \nabla_{a} Q^{\pi}(s,a) | _{a=\pi_{\theta}(s)} \right].
        \end{equation}
    \end{corollary}
    For the coherence of the content, we supplement the detailed derivation in \ref{app_derive}.
    Yet, we provide a high-level insight here. This corollary can be explained from two sides. First, the critic learned from the empirical MDP $\widehat{\mathcal{M}}$ can accurately evaluate those state-action pairs that appear in logged data $\mathcal{H}$, which makes sure the policy gradients are unbiased. This is the reason why the behavioral constraint of the actor is necessary.
    Second, the existence of regularization term $\mathcal{L}_{REG}$ in the critic essentially does not contribute to policy gradients for deterministic case, but only performs as a penalty constant that depends on whether the target policy act like the behavioral one.
    
    With this corollary at hand, we ideally inherit the simplified form of directional continuous control of actions from the DPG theorem~\cite{silver2014deterministic}.
    Then, the loss function concerning the abstracted actions $a$ is immediately acquired as
    \begin{equation}\label{eq_dpg_update}
        \mathcal{L}_{DC} = \mathbb{E}_{s \sim \mathcal{H}} - Q \left( s, \frac{\pi_{\theta}(s)}{||\pi_{\theta}(s)||_2} \right).
    \end{equation}

    \subsection{Optimization}
        
    The ultimate loss function for the optimization of the actor and critic components is summarized below to iteratively minimize
    \begin{equation}\label{eq_critic_loss}
        \mathcal{L}_{critic} = \mathcal{L}_{TD} + \alpha \cdot \mathcal{L}_{REG},
    \end{equation}
    \begin{equation}\label{eq_actor_loss}
        \mathcal{L}_{actor} = \mathcal{L}_{DC} + \beta \cdot \mathcal{L}_{BC},
    \end{equation}
    where $\alpha$ and $\beta$ are two trade-off parameters.

    Practically, to further enhance sufficiency, we basically 
    share the user preference extraction module $\mu(\cdot)$ in the actor with the critic, as suggested in \cite{xin2020self, xin2022supervised}.
    Therefore, the structure of the critic is modified to be
    \begin{equation}
        Q(s_t, a_t) = \mu(s_t) \cdot h(a_t),
    \end{equation}
    where $h(\cdot)$ is a critic-specific action feature extractor here.
    Obviously, such a design of $Q$ provides an advantage in assessing batch continuous actions $\{a_k\}_{k=1} ^K$.

    \subsection{Complexity Analysis}
    Following the notations in Section~\ref{sec_actor_complex} and the one-layer assumption of $g(\cdot)$ and $f(\cdot)$, the final model complexity of the actor is reduced from $\mathcal{O}(d|\mathcal{I}|+Tdm+m|\mathcal{I})$ to $\mathcal{O}(d|\mathcal{I}|+Tdm+md)$ with $d<<|\mathcal{I}|$, where $T$ is the upper bound of the sequence length.
    Meanwhile, the critic also enjoys the complexity reduction from $\mathcal{O}(d|\mathcal{I}|+Tdm+m|\mathcal{I})$ to $\mathcal{O}(d|\mathcal{I}|+Tdm+md+d^2)$ if $h(\cdot)$ is approximated by a one-layer perceptron.

%% file: section-V-experiments.tex
\section{Experiments}

	In this section,  we empirically examine the effectiveness of the proposed framework by investigating the following research questions.
	\begin{enumerate}[label=\textbf{RQ\arabic*}]\small
		\item Compared to existing methods, how does our framework perform in terms of capturing the complex item transition patterns, which are often evaluated by conventional top-$K$ metrics?
		
		\item As a reinforcement-learning-based framework, can ECoC achieve better outcomes in off-policy evaluation?
	
		\item Contrary to existing discrete frameworks, does our proposed continuous control method present advanced training efficiency?
		
		\item To what extent do the dual conservatism regularization terms contribute to the final performance?
	
		\item How sensitive is the performance of our framework with respect to the trade-off parameters?
	\end{enumerate}
    

\subsection{Experimental Setup}~\label{sec_exp_setup}
	\subsubsection{Datasets}
    
    We use the following three real-world datasets.
    \begin{itemize}\small
        \item \textbf{Yelp.}~\footnote{https://www.yelp.com/dataset} This dataset was collected by Yelp, which contains various information about businesses as well as abundant user profiles. However, we only focus on the sequential interactions of these users. Due to the existence of multiple versions of this dataset, we use the latest one, i.e., Yelp-2022.
        
        \item \textbf{Tmall.}~\cite{ijcai-15} This dataset is released on a TIANCHI contest sponsored by Alibaba Cloud and contains sequences of user clicks and purchases within 6 months before the ``Double 11'' campaign in 2015.
		When preparing this dataset, we retain those interactions on the same items by one user but with different feedback types for modeling the hesitation of users. 

	   \item \textbf{RL4RS.}~\cite{wang2021rl4rs} RL4RS is a real-world benchmark dataset that was recently released. It contains two main parts, i.e., RL4RS-A and RL4RS-B.
	   While the former records the user behaviors on a single page, the latter concatenates the historical behaviors across consecutive pages of one user. Ignoring the additional prediction task regarding the probability of users leaving at the end of each page, we use only RL4RS-A where the relationship between slates is not considered.
	   In addition, RL4RS provides the prices of all items, which are helpful for more precise reward signals.

    \end{itemize}
    

\begin{table}[ht]
	\centering\footnotesize
	\caption{Dataset Statistics}\label{tab_data}%
	\begin{tabular}{lcccccc}
	\toprule
	\multirow{3}{*}{Dataset} & \multicolumn{2}{c}{Filter Param. (minimal)} & \multicolumn{3}{c}{After Preprocessing} & \multirow{3}{*}{Sparsity} \\
	\cmidrule(lr){2-3} \cmidrule(lr){4-6} & \makecell{\#support\\ of item} & \makecell{\#length \\ of session} & \#users & \#items & \#interactions \\
	\midrule
	Yelp  & 10  & 10  & 72,487  &  43,748  &  2,043,402 & 0.9994 \\
	Tmall & 20  & 10 & 238,242 & 113,183 & 9,795,393 & 0.9996 \\
	RL4RS-A &  /  &  /  & 149,414 & 283  & 15,473,844 & 0.6341 \\
	\bottomrule
	\end{tabular}%
\end{table}%
	
    After preprocessing, some basic statistics are listed in Table~\ref{tab_data}. 
    Immediately, we can find that with much fewer candidate items contained, the sparsity of RL4RS-A is much lower. As a result, with more frequent interaction attempts on single items in addition to the positive feedback from users, this dataset is more suitable for testing RL-based recommender systems~\cite{wang2021rl4rs}.
    However, it is also necessary to test the efficiency and effectiveness on datasets like Yelp and Tmall, which have numerous items.
    This is the reason why we select datasets that cover the different levels of the scale of items contained.

	For the splitting of the whole datasets, we use 80\% interactions for training and leave the remaining for testing. Note that such an operation follows the timestamps recorded. As the only exception, the separation of RL4RS-A is conducted by executing the accompanying scripts. In addition, we also randomly select 10\% from the training set as the validation set for tuning trade-off parameters.

	\subsubsection{Evaluation Metrics}~\label{sec_eval_metric}
    For a comprehensive understanding of the performance of our model as well as the baselines, we test from two main aspects, where the evaluation criteria are accordingly selected.
	
	First, to assess the ability to recover behavioral policy from logged data, metrics including top-$k$ Recall (a.k.a. Hit Ratio \textbf{HR@$k$}), Mean Reciprocal Rank (\textbf{MRR@$k$}), and Normalized Discounted Cumulative Gain (\textbf{NDCG@$k$}) are widely adopted in previous studies~\cite{tang2018personalized, kang2018self, wu2019session}.
	HR@$k$ measures the ratio of the relevant items in the recommended list, while MRR@$k$ and NDCG@$k$ further take into account the ranking order and assign higher scores to relevant items ranked higher.
	For the complete comparisons in Section~\ref{sec_exp_imit}, we display these metrics when $k$ is 5,10 and 20. While for the rest experiments, we only report the results with $k=20$ as results for other $k$ have similar tendencies.

\begin{figure}[ht]
    \centering
    \includegraphics[width=0.65\columnwidth]{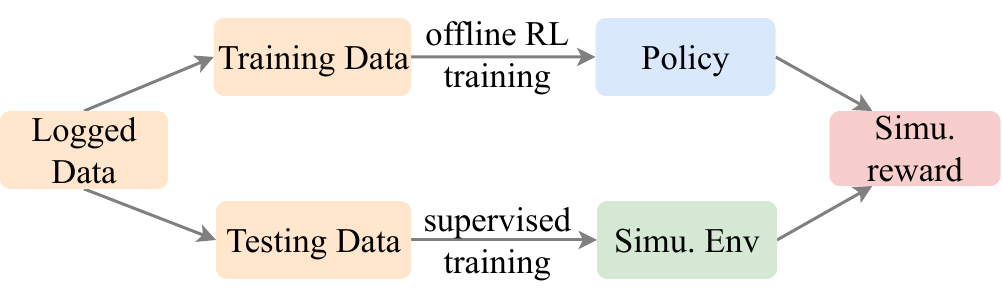}
    \caption{The workflow of off-policy evaluation.}
    \label{fig_eval_flow}
\end{figure}
	
	Besides imitation performance, another concerned metric is the off-policy evaluation (OPE) of learned policies, especially when online deployment is infeasible~\cite{wang2021rl4rs}.
	Following the benchmark of RL4RS, the OPE of the policies from training data is assessed by interacting with the simulated environment fitted from the testing data and recording the total rewards. The entire workflow is shown in Figure~\ref{fig_eval_flow}.
	
	In particular, we select different reward settings in terms of the capability of different datasets.
	For Yelp, the explicit discrete scores ranging from 1 to 5 can be utilized to indicate whether the user recognizes the recommended items. While on Tmall, we instead assign different values for different kinds of feedback. For example, the reward is 1 if the recommended item is clicked, 2 if the item is favored and 3 if a purchase is achieved. Lastly, we further leverage the price of items regarding the rewards on RL4RS-A. If the recommended item is finally bought by the user, the exact reward is the price otherwise 0.
    
	\subsubsection{Baseline Methods}

    To comprehensively perceive the performance of our ECoC framework, we select a wide range of existing RL-based sequential recommendation methods, along with one state-of-the-art bandit algorithm. We list them as follows.
    \begin{itemize}\small
		\item \textbf{NeuralLCB}~\cite{nguyen-tang2022offline} --- Neural Lower Confidence Bound considers the general offline contextual bandit problems with neural networks as function approximation. We apply this method for the sequential recommendation.
    	
		\item \textbf{SQN}~\cite{xin2020self} --- Self-supervised Q-learning is the first offline RL learning method for sequential recommendation, which combines supervised learning with classic Q-learning through a shared backbone supervised model.
    	
		\item \textbf{SDAC+SR}/\textbf{DC}~\cite{xiao2021general} --- Stochastic discrete actor-critic is a strong state-of-the-art framework for discrete interactive recommendation scenarios. With the probabilistic formulation, several constraint approaches are proposed to solve the extrapolation error. We select 2 strong ones among them, including SDAC with supervised regularization (SDAC+SR) and dual constraints (SDAC+DC).
    	
		\item \textbf{PRL}~\cite{xin2022rethinking} --- Prompt-based reinforcement learning proposes a new paradigm for RL-based recommender systems in directly predicting the actions given the state-reward pairs.
    	
		\item \textbf{SA2C}~\cite{xin2022supervised} --- Supervised advantage actor-critic extends SQN in the way of re-weighting the logits outputted by the actor by the Q values from the critic.
    \end{itemize}
    Again, we emphasize that these methods are proposed for discrete action space.
    In addition, it should be noted that each of these baselines, except for NeuralLCB, are implemented with three different supervised models as the backbone, i.e.,
    \begin{itemize}\small
    	\item \textbf{NextItNet}~\cite{yuan2019simple} --- This method uses a dilated CNN to enlarge the receptive fields in order to learn from long user behavior sequences. Moreover, the adopted residual connections help increase the network depth. Overall, NextItNet achieves good performance with high efficiency.

		\item \textbf{SASRec}~\cite{tang2018personalized} --- Self-attention-based sequential model applies the Transformer structure to capture the long-term semantics, regardless of whether the interactions are sparse or dense.

    	\item \textbf{SR-GNN}~\cite{wu2019session} --- This method first regards the item transitions as graph-structure data and leverages GNN propagations to capture the complex transition patterns among items.
    \end{itemize}
	These backbones might not be the most sophisticated ones among their own classifications. However, what we really care about is the robustness and stability of the proposed ECoC framework. Hence, it is the coverage of various network structures, including CNNs, GNNs and Transformers, that indeed matters.
    

	\subsubsection{Implementation Details}

    All models are implemented by PyTorch~\cite{paszke2019pytorch} and run
    5 times with different random seeds on a single NVIDIA GeForce RTX 3090 GPU, with the averaged values reported.
    In general, the length of historical sequences on Yelp and Tmall is truncated to a maximum of 50 in terms of the most recent behaviors. Whereas on RL4RS-A, the number is 8. Sequences of which the length is less than the number are padded by a padding item.
    All parameters are trained by the Adam optimizer, with a learning rate of $10^{-3}$ and an $L_2$ penalty of $10^{-5}$. Moreover, the batch size is chosen as 256 for all datasets. The embedding dimension $d$ is set as 64 for Yelp and Tmall and 32 for RL4RS-A. The dropout ratio for embedding vectors is fixed as 0.2.
    The negative samples in $\mathcal{L}_{BC}$ are uniformly sampled, while the number $N_2$ is $10^4$ on Yelp and Tmall. With fewer items contained in RL4RS-A, we directly replace the BPR loss with the cross-entropy loss.

    The details of the backbones we selected are as follows.
    For NextItNet,  we utilize the design of residual block (b) and stack two blocks with dilation factor as [1, 2, 4, 8] on Yelp and Tmall. While on RL4RS-A, the dilation factor is reduced to be [1, 2, 4]. For SASRec, the numbers of blocks and heads of self-attention in each block are set as 1 and 2 on Yelp and Tmall, respectively. On RL4RS-A, the numbers are increased to 2 and 4 correspondingly.
    For SR-GNN, the numbers of propagation steps are chosen as 1, 1, and 3 on Yelp, Tmall, and RL4RS-A separately.
    
    Last but not least, we introduce the RL-related settings. Following \cite{xin2022supervised}, the discount factor $\gamma$ in all RL methods is set as 0.5. The implementations of SQN, PRL and SA2C follow their original papers, except for the reward settings. All RL methods are conducted by the same reward settings illustrated in Section~\ref{sec_eval_metric} for a fair comparison. For our ECoC, the trade-off parameter $\alpha$ is tuned as 5, 5, and 8 on Yelp, Tmall and RL4RS-A, whereas the values of $\beta$ are always 1.
    The number of negative sampling $N_1$ is tuned as 500 on Yelp and Tmall. The influences of these parameters are further discussed in the experimental part.

\begin{table}
	\centering\footnotesize
	\setlength\tabcolsep{4pt}
	\caption{Imitation performance and training time per epoch of methods on Yelp with different backbones. The results of the best-performing baseline are underlined. 
    }\label{tab_imit_yelp} 
	\begin{tabular}{cl||cc|cc|cc|c}
		\toprule
		& \multicolumn{1}{c}{Method} & HR@5 & HR@10 & MRR@5 & MRR@10 & NG@5 & NG@10 & Time(s) \\
		\midrule
		& NeuralLCB & .0258 & .0455 & .0128 & .0149 & .0160 & .0219 & 80.3 \\
		\midrule
		\multirow{7}{*}{\rotatebox[origin=c]{90}{NextItNet}} & Backbone & .0225 & .0396 & .011 & .0132 & .0138 & .0193 & 56.1 \\
		& SQN & .0235 & .0416 & .0113 & .0135 & .0144 & .0200 & 288.4 \\
		& PRL & .0238 & .0425 & .0115 & .0139 & .0145 & .0205 & 346.9 \\
		& SDAC+SR & \underline{.0246} & \underline{.0436} & \underline{.0117} & .0141 & .0149 & .0208 & 449.3 \\
		& SDAC+DC & .0237 & .0421 & .0113 & .0138 & .0144 & .0203 & 457.5 \\
		& SA2C & .0240 & .0432 & \underline{.0117} & \underline{.0142} & \underline{.0150} & \underline{.0210} & 325.4 \\
		\cmidrule(lr){2-9}
		& ECoC & \textbf{.0262} & \textbf{.0467} & \textbf{.0125} & \textbf{.0152} & \textbf{.0159} & \textbf{.0224} & 218.2 \\
		\midrule
		\multirow{7}{*}{\rotatebox[origin=c]{90}{SASRec}} & Backbone & .0263 & .0466 & .0126 & .0152 & .0161 & .0228 & 42.8 \\
		& SQN & .0270 & .0483 & .0130 & .0158 & .0164 & .0232 & 207.7 \\
		& PRL & .0285 & .0502 & .0137 & .0165 & .0175 & .0245 & 282.8 \\
		& SDAC+SR & \underline{.0290} & \underline{.0510} & \underline{.0142} & \underline{.0172} & \underline{.0180} & \underline{.0252} & 350.5 \\
		& SDAC+DC & .0282 & .0495 & .0137 & .0164 & .0175 & .0242 & 361.3 \\
		& SA2C & .0286 & .0506 & .0139 & .0168 & .0178 & .0248 & 233.8 \\
		\cmidrule(lr){2-9}
		& ECoC & \textbf{.0317} & \textbf{.0553} & \textbf{.0154} & \textbf{.0185} & \textbf{.0196} & \textbf{.0275} & 150.7 \\
		\midrule
		\multirow{7}{*}{\rotatebox[origin=c]{90}{SR-GNN}} & Backbone & .0239 & .0430 & .0114 & .0139 & .0145 & .0206 & 136.3 \\
		& SQN & .0257 & .0453 & .0.124 & .0147 & .0154 & .0217 & 735.7 \\
		& PRL & .0259 & .0461 & .0122 & .0147 & .0156 & .0222 & 903.6 \\
		& SDAC+SR & .0262 & .0464 & .0126 & .0151 & .0158 & .0223 & 1215.7 \\
		& SDAC+DC & .0245 & .0441 & .0119 & .0141 & .0149 & .0210 & 1237.4 \\
		& SA2C & \underline{.0268} & \underline{.0476} & \underline{.0130} & \underline{.0158} & \underline{.0165} & \underline{.0231} & 820.1 \\
		\cmidrule(lr){2-9}
		& ECoC & \textbf{.0290} & \textbf{.0519} & \textbf{.0143} & \textbf{.0173} & \textbf{.0179} & \textbf{.0253} & 547.3 \\
		\bottomrule
	\end{tabular}%
\end{table}%

\begin{table}
	\centering\footnotesize
	\setlength\tabcolsep{4pt}
	\caption{Imitation performance and training time per epoch of methods on Tmall with different backbones. The results of the best-performing baseline are underlined. 
    }\label{tab_imit_tmall}
	\begin{tabular}{cl||cc|cc|cc|c}
		\toprule
		& \multicolumn{1}{c}{Method} & HR@5 & HR@10 & MRR@5 & MRR@10 & NG@5 & NG@10 & Time(s)\\
		\midrule
		& NeuralLCB & .1621 & .2183 & .1021 & .1093 & .1170 & .1349 & 296.7 \\
		\midrule
		\multirow{7}{*}{\rotatebox[origin=c]{90}{NextItNet}} & Backbone & .1296 & .1775 & .0763 & .0827 & .0895 & .1050 & 214.8 \\
		& SQN & .1410 & .1905 & .0848 & .0914 & .0987 & .1147 & 1029.1 \\
		& PRL & .1559 & .2085 & .0947 & .1017 & .1099 & .1269 & 1325.5 \\
		& SDAC+SR & \underline{.1718} & \underline{.2274} & \underline{.1055} & \underline{.1129} & \underline{.1219} & \underline{.1399} & 1867.8 \\
		& SDAC+DC & .1426 & .1932 & .0855 & .0923 & .0997 & .1160 & 1893.4 \\
		& SA2C & .1653 & .2208 & .1032 & .1109 & .1189 & .1368 & 1178.4 \\
		\cmidrule(lr){2-9}
		& ECoC & \textbf{.1861} & \textbf{.2471} & \textbf{.1144} & \textbf{.1225} & \textbf{.1322} & \textbf{.1519} & 663.8 \\
		\midrule
		\multirow{7}{*}{\rotatebox[origin=c]{90}{SASRec}} & Backbone & .1567 & .2095 & .0954 & .1024 & .1113 & .1284 & 133.3 \\
		& SQN & .1668 & .2221 & .1024 & .1098 & .1184 & .1364 & 670.5 \\
		& PRL & .1798 & .2397 & .1082 & .1161 & .1253 & .1445 & 869.2 \\
		& SDAC+SR & .1822 & .2419 & .1111 & .1191 & .1286 & .1480 & 1095.6 \\
		& SDAC+DC & .1753 & .2351 & .1070 & .1150 & .1240 & .1430 & 1110.5 \\
		& SA2C & \underline{.1828} & \underline{.2435} & \underline{.1113} & \underline{.1194} & \underline{.1291} & \underline{.1487} & 782.4 \\
		\cmidrule(lr){2-9}
		& ECoC & \textbf{.1990} & \textbf{.2620} & \textbf{.1234} & \textbf{.1322} & \textbf{.1420} & \textbf{.1628} & 438.3 \\
		\midrule
		\multirow{7}{*}{\rotatebox[origin=c]{90}{SR-GNN}} & Backbone & .1742 & .2311 & .1081 & .1148 & .1243 & .1423 & 460.9 \\
		& SQN & .1854 & .2434 & .1157 & .1235 & .1330 & .1518 & 2148.8 \\
		& PRL & .1905 & .2505 & .1191 & .1270 & .1369 & .1562 & 2568.1 \\
		& SDAC+SR & \underline{.1948} & \underline{.2567} & \underline{.1205} & \underline{.1287} & \underline{.1385} & \underline{.1587} & 3832.5 \\
		& SDAC+DC & .1838 & .2496 & .1165 & .1246 & .1346 & .1543 & 3902.2 \\
		& SA2C & .1895 & .2557 & .1201 & .1285 & .1375 & .1574 & 2356.4 \\
		\cmidrule(lr){2-9}
		& ECoC & \textbf{.2149} & \textbf{.2796} & \textbf{.1334} & \textbf{.1421} & \textbf{.1536} & \textbf{.1745} &  1370.2 \\
		\bottomrule
	\end{tabular}%
\end{table}%

\begin{table}
	\centering\footnotesize
	\setlength\tabcolsep{4pt}
	\caption{Imitation performance and training time per epoch of methods on RL4RS-A with different backbones. The results of the best-performing baseline are underlined. 
    }\label{tab_imit_rl4rs}
	\begin{tabular}{cl||cc|cc|cc|c}
		\toprule
		& \multicolumn{1}{c}{Method} & HR@5 & HR@10 & MRR@5 & MRR@10 & NG@5 & NG@10 & Time(s) \\
		\midrule
		& NeuralLCB & .4834 & .6449 & .3053 & .3267 & .3496 & .4016 & 30.2 \\
		\midrule
		\multirow{7}{*}{\rotatebox[origin=c]{90}{NextItNet}} & Backbone  & .4886 & .6487 & .3090 & .3301 & .3536 & .4051 & 31.5 \\
		& SQN & .4924 & .6519 & .3122 & .3332 & .3570 & .4083 & 68.4 \\
		& PRL & .4976 & .6578 & .3151 & .3363 & .3604 & .4121 & 97.9 \\
		& SDAC+SR & \underline{.5006} & \underline{.6614} & \underline{.3170} & \underline{.3383} & \underline{.3626} & \underline{.4144} & 113.3 \\
		& SDAC+DC & .4932 & .6531 & .3126 & .3337 & .3574 & .4089 & 117.5 \\
		& SA2C & .4992 & .6577 & .3168 & .3378 & .3621 & .4132 & 81.4 \\
		\cmidrule(lr){2-9}
		& ECoC & \textbf{.5074} & \textbf{.6694} & \textbf{.3217} & \textbf{.3432} & \textbf{.3678} & \textbf{.4200} & 62.4 \\
		\midrule
		\multirow{7}{*}{\rotatebox[origin=c]{90}{SASRec}} & Backbone & .4805 & .6406 & .3021 & .3232 & .3464 & .3980 & 28.4 \\
		& SQN & .4825 & .6445 & .3044 & .3259 & .3487 & .4008 & 58.5 \\
		& PRL & .4871 & .6487 & .3076 & .3290 & .3522 & .4043 & 75.6 \\
		& SDAC+SR & \underline{.4893} & \underline{.6509} & \underline{.3089} & \underline{.3303} & \underline{.3537} & \underline{.4058} & 92.1\\
		& SDAC+DC & .4846 & .6464 & .3048 & .3262 & .3495 & .4016 & 96.9 \\
		& SA2C & .4877 & .6496 & .3079 & .3293 & .3526 & .4047 & 67.1 \\
		\cmidrule(lr){2-9}
		& ECoC & \textbf{.4980} & \textbf{.6579} & \textbf{.3153} & \textbf{.3365} & \textbf{.3607} & \textbf{.4122} & 51.7 \\
		\midrule
		\multirow{7}{*}{\rotatebox[origin=c]{90}{SR-GNN}} & Backbone & .4914 & .6521 & .3113 & .3325 & .3560 & .4077 & 82.9 \\
		& SQN & .5015 & .6621 & .3175 & .3388 & .3632 & .4150 & 176.4 \\
		& PRL & .5039 & .6646 & .3202 & .3415 & .3658 & .4176 & 211.2 \\
		& SDAC+SR & \underline{.5060} & .6656 & \underline{.3213} & \underline{.3425} & \underline{.3672} & .4187 & 266.3 \\
		& SDAC+DC & .4990 & .6598 & .3155 & .3368 & .3611 & .4129 & 275.1 \\
		& SA2C & .5051 & \underline{.6665} & .3201 & .3417 & .3663 & \underline{.4188} & 194.7 \\
		\cmidrule(lr){2-9}
		& ECoC & \textbf{.5156} & \textbf{.6784} & \textbf{.3268} & \textbf{.3484} & \textbf{.3737} & \textbf{.4262} & 159.1 \\
		\bottomrule
	\end{tabular}%
\end{table}%

\subsection{Experimental Results}

	\subsubsection{Performance Comparisons\ \textbf{(RQ1)}}\label{sec_exp_imit}

	Table~\ref{tab_imit_yelp}, Table~\ref{tab_imit_tmall} and Table~\ref{tab_imit_rl4rs} display the imitation performance of various RL-based methods on three datasets.
	Overall,  our framework ECoC achieves consistently better performance over baseline methods when equipped with different supervised methods as backbones.
	Generally, ECoC achieves an average increase of 8.25\%, 9.20\% and 1.46\% respectively on Yelp, Tmall and RL4RS-A, compared with the best results on previous methods.
	These results uncover that effectively managing a fine-grained control of continuous representations of user preference is beneficial to improve the capability of predicting the historical item transitions in logged data.

	However, the very first observation is that the improvement ratios seem to be marginal on RL4RS-A compared to those on other datasets. This could be attributed to two potential reasons. On one hand, with extremely fewer items (i.e., 283) and much denser interactions, acquiring a satisfactory recommendation outcome on this dataset is fairly easy, which can be confirmed by the metrics. On the other hand, the power of unified evaluation in policy evaluation is limited without sufficient items nearby, especially in high-dimensional continuous action spaces.

	Secondly, for the performance of three different backbones, our ECoC and baseline methods with SR-GNN as backbones show a relatively stronger ability to encode the behavior sequences and infer user interest. This may be because GNN is more suitable to explore complex graph-structured data, as recently validated in \cite{zhang2023efficiently}.
	However, no matter what the performance of backbones is, of particular importance is the consistent improvement of our ECoC, which verifies its stability and generalization among various network structures.

	Thirdly, all RL-based models outperform those supervised backbones, which suggests properly considering the reward signals can enhance the ability of recommender systems to capture the item transitions. Though the ultimate goal of RL-based recommendation agents is to activate long-term user engagement, such precise transition modeling still plays a key role in decision optimization. 

	Fourthly, we would highlight that the sparsity of Yelp and Tmall does not affect the dominant performance of ECoC, which is exactly the value of our continuous control framework. It might reveal that the policy is well-guided to determine the user preference closer to the region where the next item is most likely to appear. This could be attributed to the proposed unified evaluation technique, which directly takes advantage of those particular item representations when evaluating the actions.
	

\begin{figure}[ht]
	\centering
	\begin{subfigure}[ht]{0.32\columnwidth}
		\includegraphics[width=1.\textwidth]{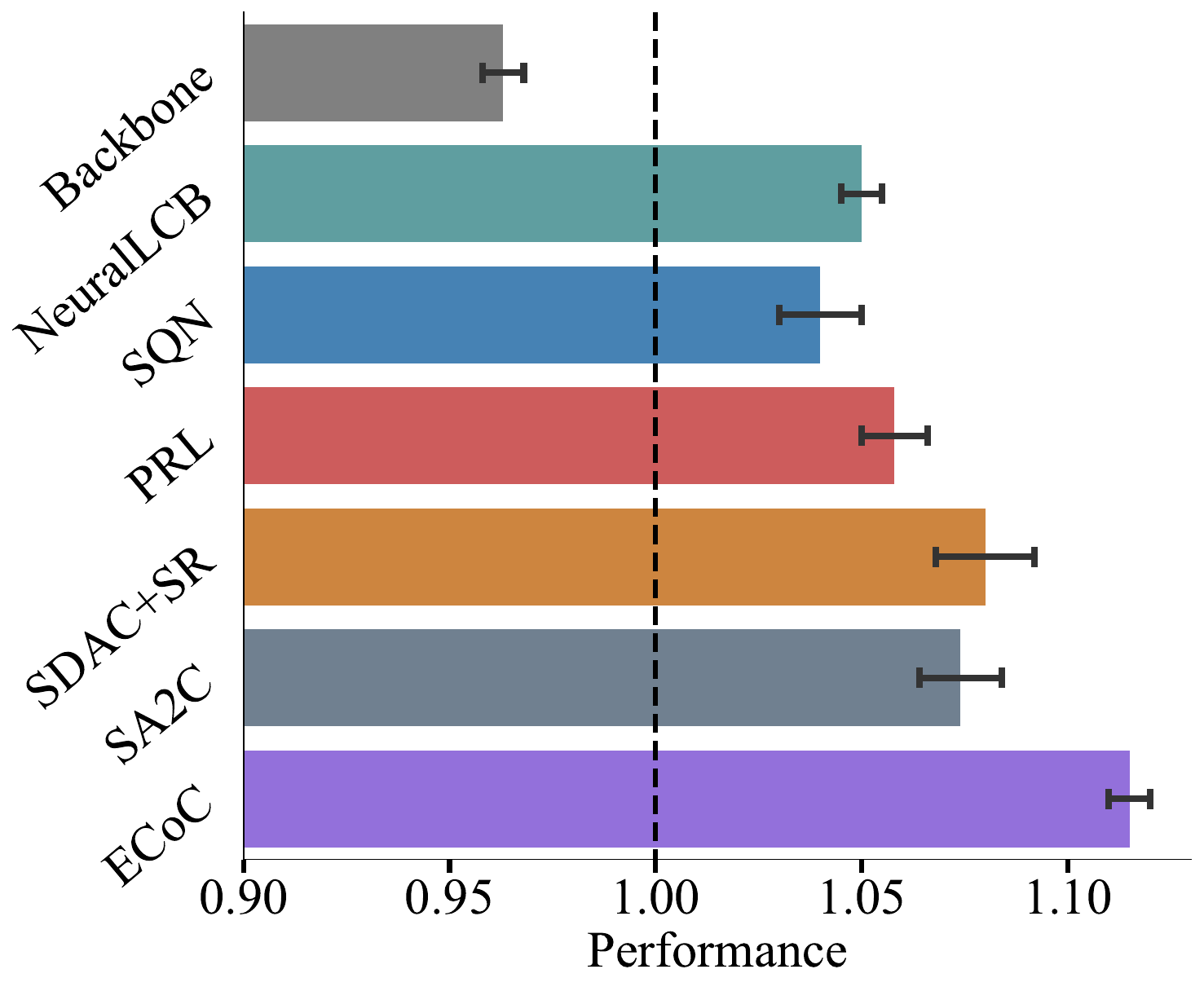}
		\subcaption{Yelp}
	\end{subfigure}
	\hfill
	\begin{subfigure}[ht]{0.32\columnwidth}
		\includegraphics[width=1.\textwidth]{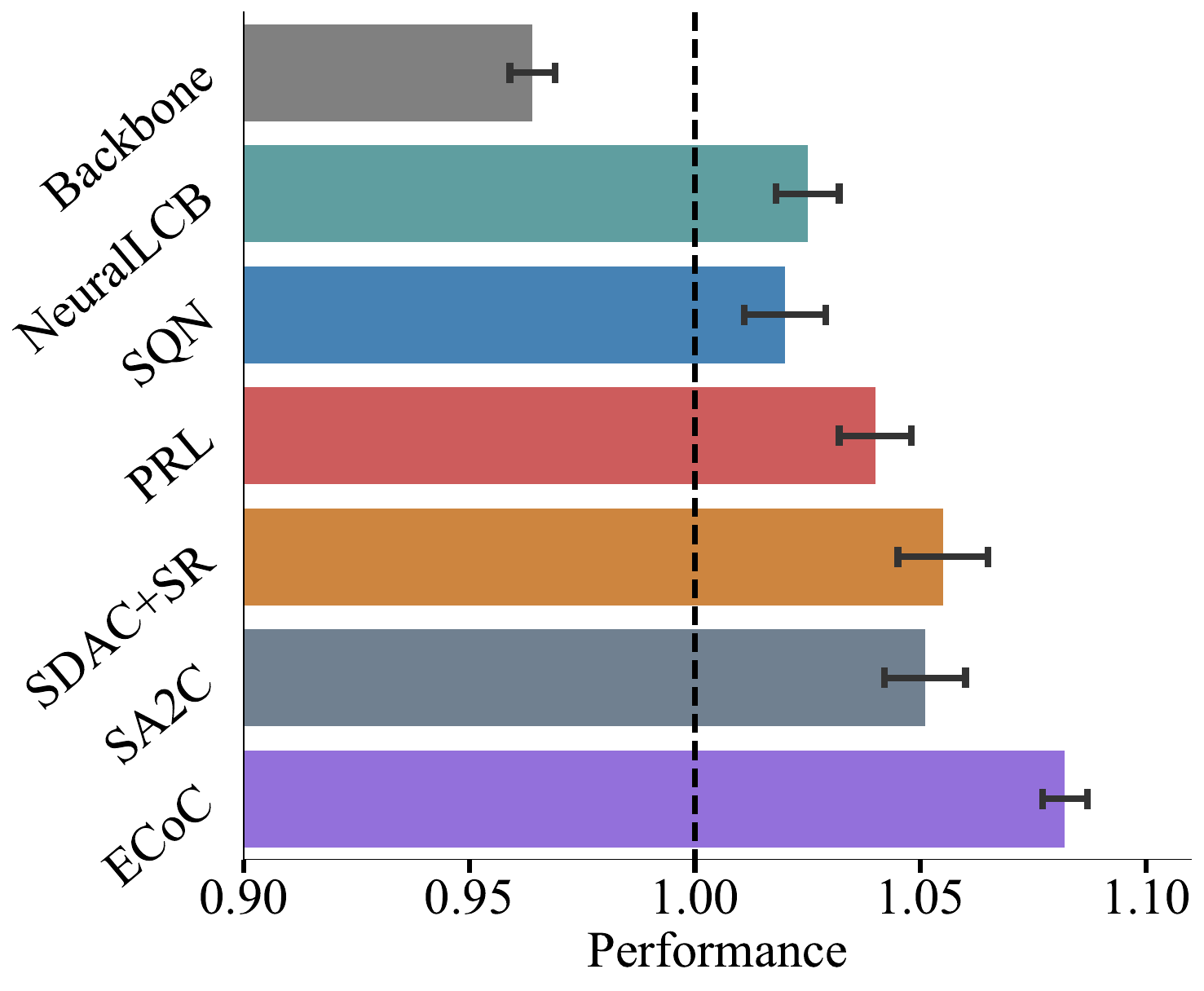}
		\subcaption{Tmall}
	\end{subfigure}
	\hfill
	\begin{subfigure}[ht]{0.32\columnwidth}
		\includegraphics[width=1.\textwidth]{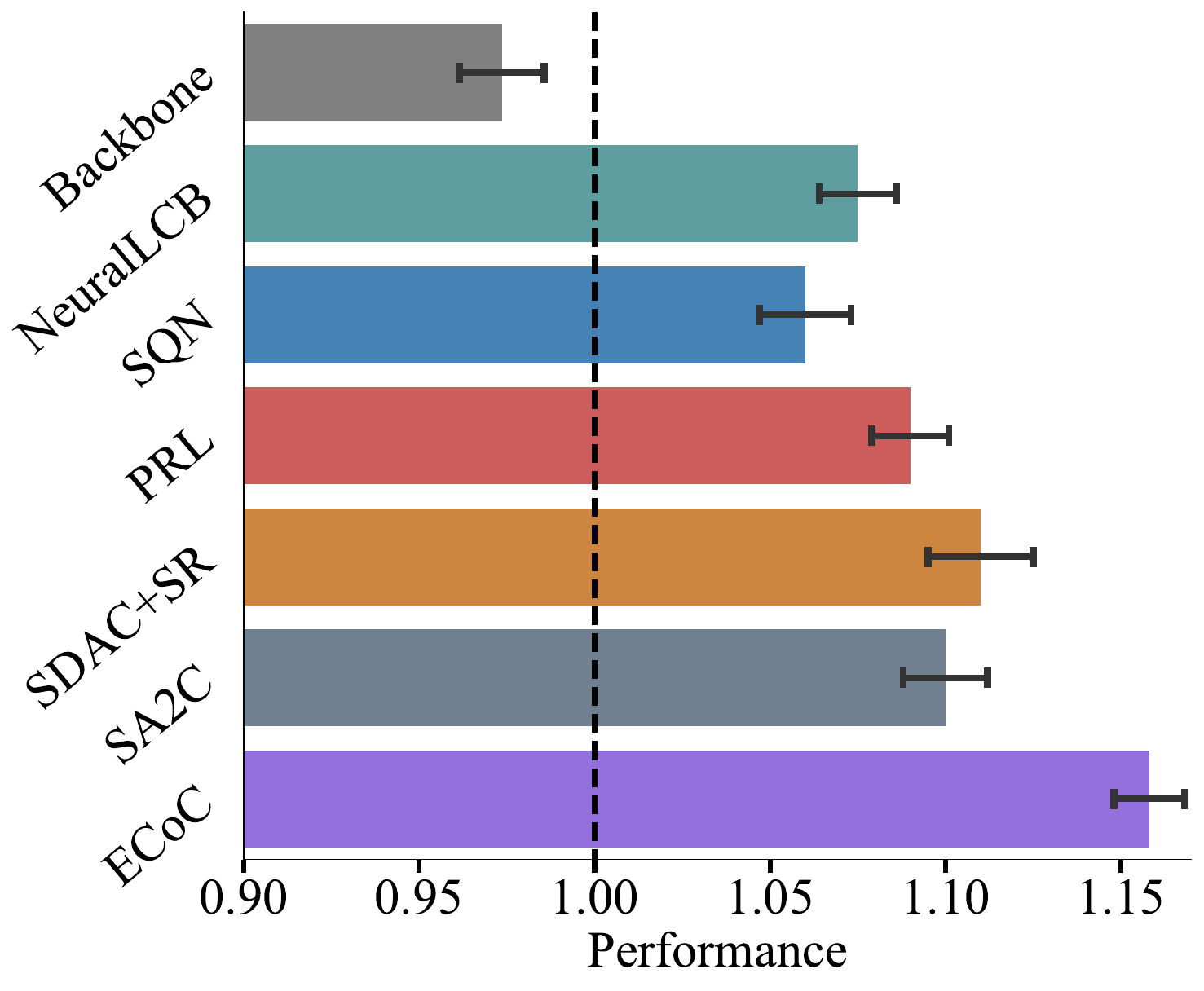}
		\subcaption{RL4RL-A}
	\end{subfigure}
	\caption{Performance ratios of total rewards gained from the simulated environment built by testing data. The baseline value (i.e., 1) represents the performance of logged data. All methods are implemented with SR-GNN as the backbone.}
	\label{fig_ope}
\end{figure}

	\subsubsection{Reward Investigation\ \textbf{(RQ2)}}\label{sec_exp_ope}

	In this subsection, we focus on the investigation of off-policy performance, which is a key indicator for RL-based recommender systems.
	As illustrated in Section~\ref{sec_eval_metric}, we test the predefined reward values from the top-1 position recommendation outcomes of different algorithms. Due to the lack of terminal prediction modules, we directly leverage those historical sequences in testing data and report the total rewards all methods can get, where the virtual feedback is generated by the simulated environment. Note that all results are normalized by the baseline performance, which is the reward value computed from the logged actions.

    From the comparisons in Figure~\ref{fig_ope}, we observe that ECoC also performs the best in taking the most rewarding actions. 
    Moreover, contrary to the backbone, the better off-policy performance of bandit and RL-based algorithms over logged data suggests an excellent advantage in optimizing the long-term utility.
    While for the improvement proportions over baselines among three datasets, it is obviously easiest for RL-based recommender systems to gain a significant boost on RL4RS-A than Yelp and Tmall due to the denser interactions among fewer items.

	\subsubsection{Training Efficiency\ \textbf{(RQ3)}}

    The empirical training efficiency of different RL-based recommender systems is investigated in this subsection.
	We have listed the running time on GPUs per epoch for each model along with the imitation performance in Table~\ref{tab_imit_yelp}, Table~\ref{tab_imit_tmall} and Table~\ref{tab_imit_rl4rs}.
    Unsurprisingly, beneficial from continuous implementation, our ECoC spends averaged 25.8\%, 35.4\% and 10.1\% less time on Yelp, Tmall and RL4RS-A respectively, compared with the best performing discrete reinforcement learning frameworks. More importantly, as the scale of items contained in datasets increases, the improvement ratios on training efficiency tend to grow as well, which confirms the potential of our ECoC method.

\begin{figure}[ht]
	\centering
	\begin{subfigure}[ht]{0.32\textwidth}
		\includegraphics[width=1.\textwidth]{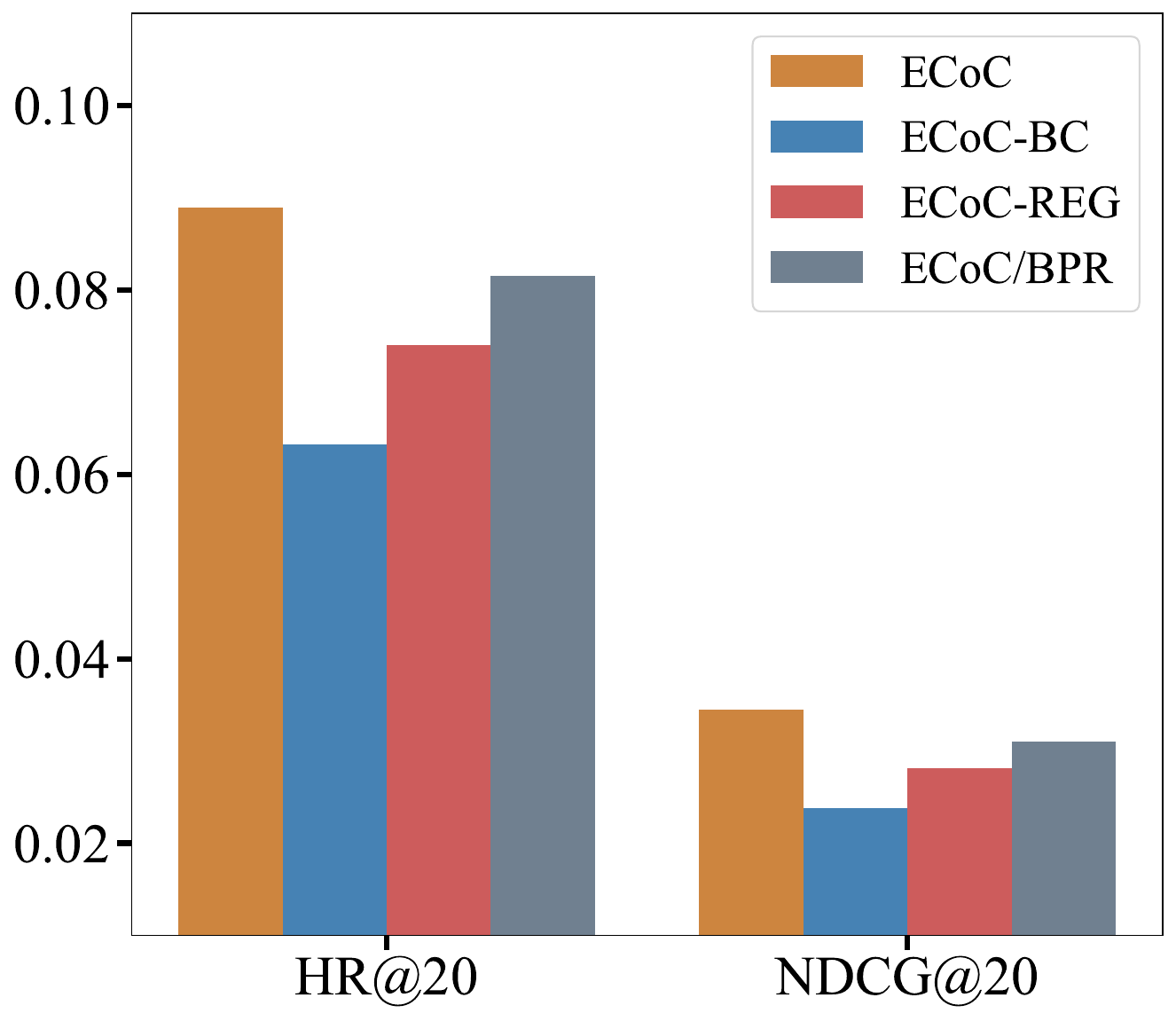}
		\subcaption{Yelp}
	\end{subfigure}
	\hfill
	\begin{subfigure}[ht]{0.32\textwidth}
		\includegraphics[width=1.\textwidth]{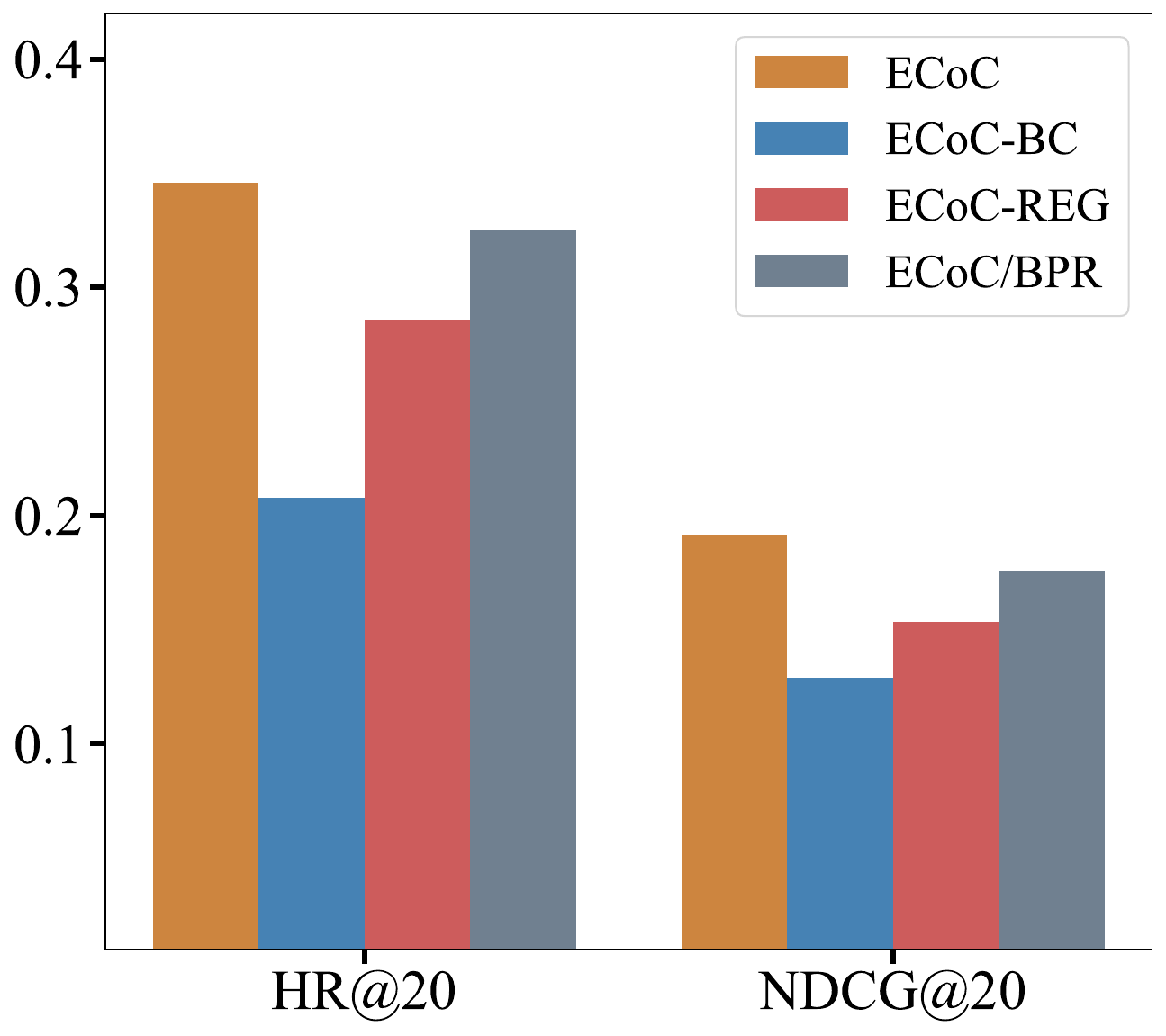}
		\subcaption{Tmall}
	\end{subfigure}
	\hfill
	\begin{subfigure}[ht]{0.32\textwidth}
		\includegraphics[width=1.\textwidth]{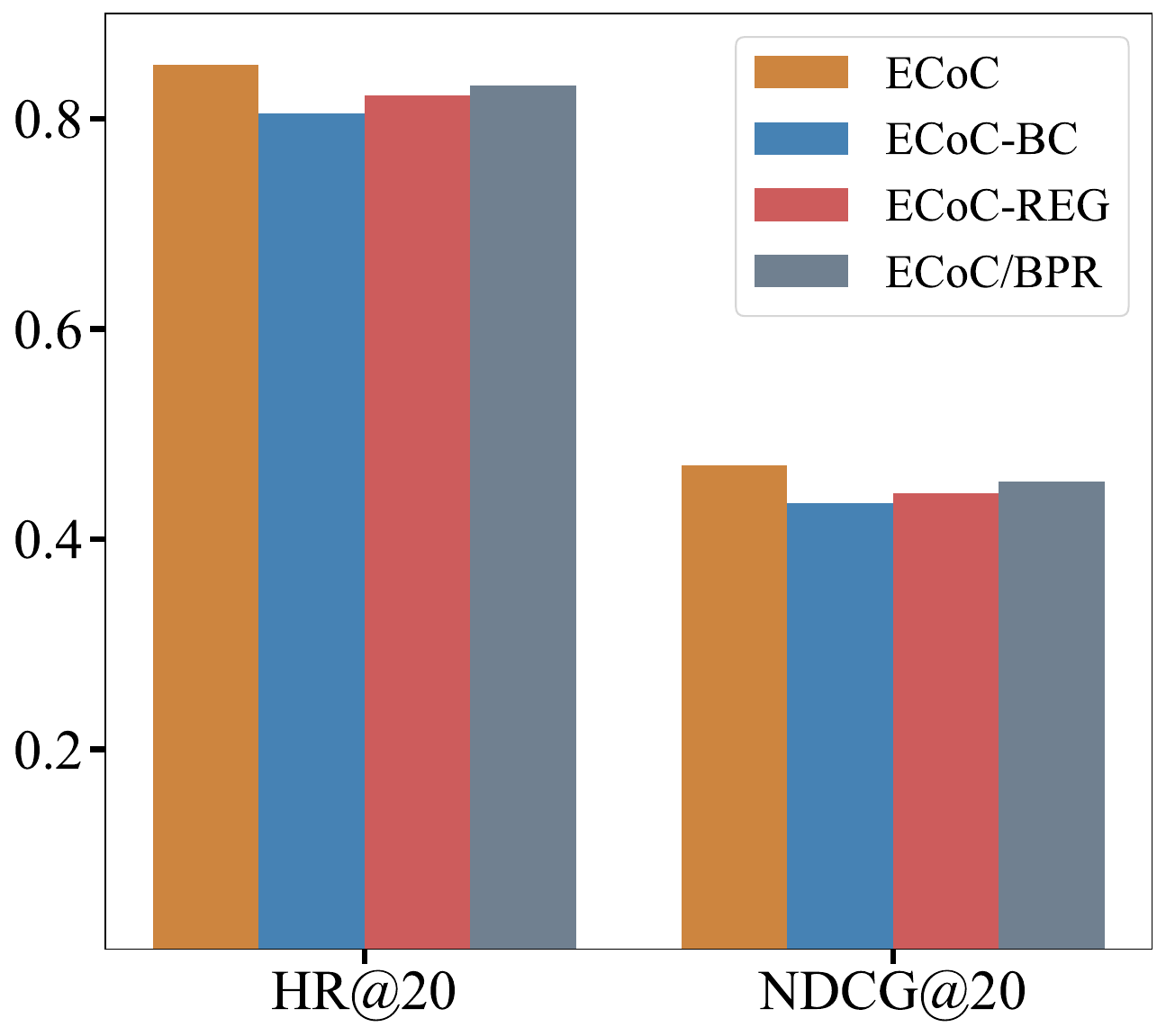}
		\subcaption{RL4RL-A}
	\end{subfigure}
	\caption{Comparison of Variants of ECoC with SR-GNN as the backbone on three datasets.}
	\label{fig_ablation}
\end{figure}

	\subsubsection{Ablation Study\ \textbf{(RQ4)}}\label{sec_ablation}
	
    In this subsection, we dive deep into the functional analysis of different loss terms.
	As illustrated in Figure~\ref{fig_ablation}, we compare the results of ECoC without the behavioral constraint in the actor component $\mathcal{L}_{BC}$ (i.e., ECoC-BC), ECoC without the conservatism regularization $\mathcal{L}_{REG}$ in the critic component (i.e., ECoC-REG) and ECoC with $\mathcal{L}_{REG}$ replaced by the BPR loss (i.e., ECoC/BPR).
    
	First, it is evident that ECoC-BC encounters the sharpest performance drop due to the lack of a behavioral constraint.
	This finding is significant since under the end-to-end learning frameworks, such a constraint might not only act as a regularization term, but also be beneficial for the representation learning of the whole item space (i.e., $\mathrm{M}_{\mathcal{I}}$), which is also indicated in \cite{liu2020end}.
    Secondly, without conservatism guarantee $\mathcal{L}_{REG}$ in the process of policy evaluation, it is hard for the critic to specify the correct update directions for the actor, leading the policy learned by ECoC-REG to perform worse than the backbone.
    Thirdly, we are also interested in whether the conventional BPR loss can cooperate well with the training of the critic. As a result, the variant ECoC/BPR exhibits slightly declined results, which reveals that ECoC actually learns a tighter conservatism upper bound with $\mathcal{L}_{REG}$.

\begin{figure}[ht]
	\centering
	\begin{subfigure}[ht]{0.32\textwidth}
		\includegraphics[width=1.\textwidth]{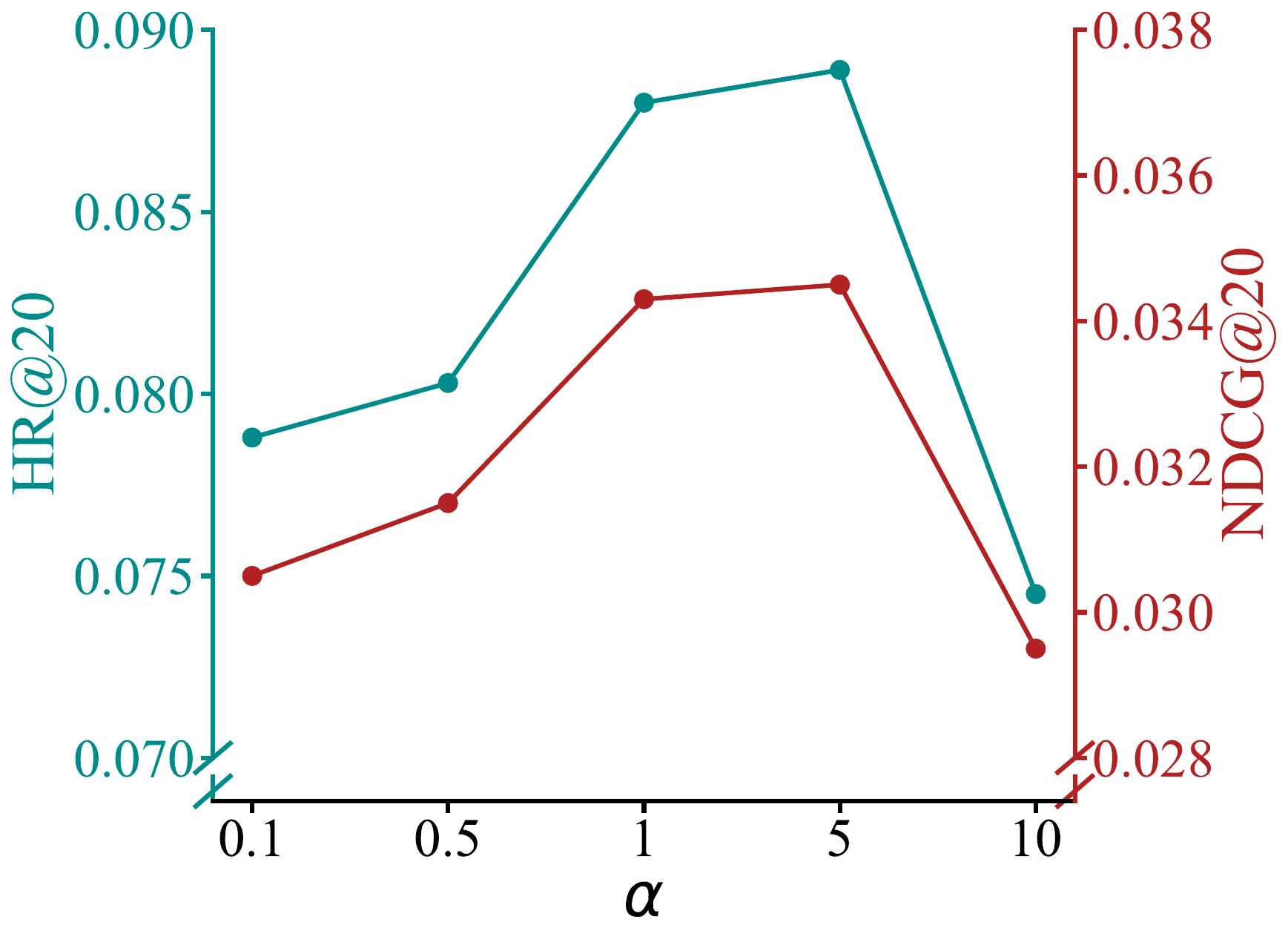}
		\subcaption{Yelp}
	\end{subfigure}
	\hfill
	\begin{subfigure}[ht]{0.32\textwidth}
		\includegraphics[width=1.\textwidth]{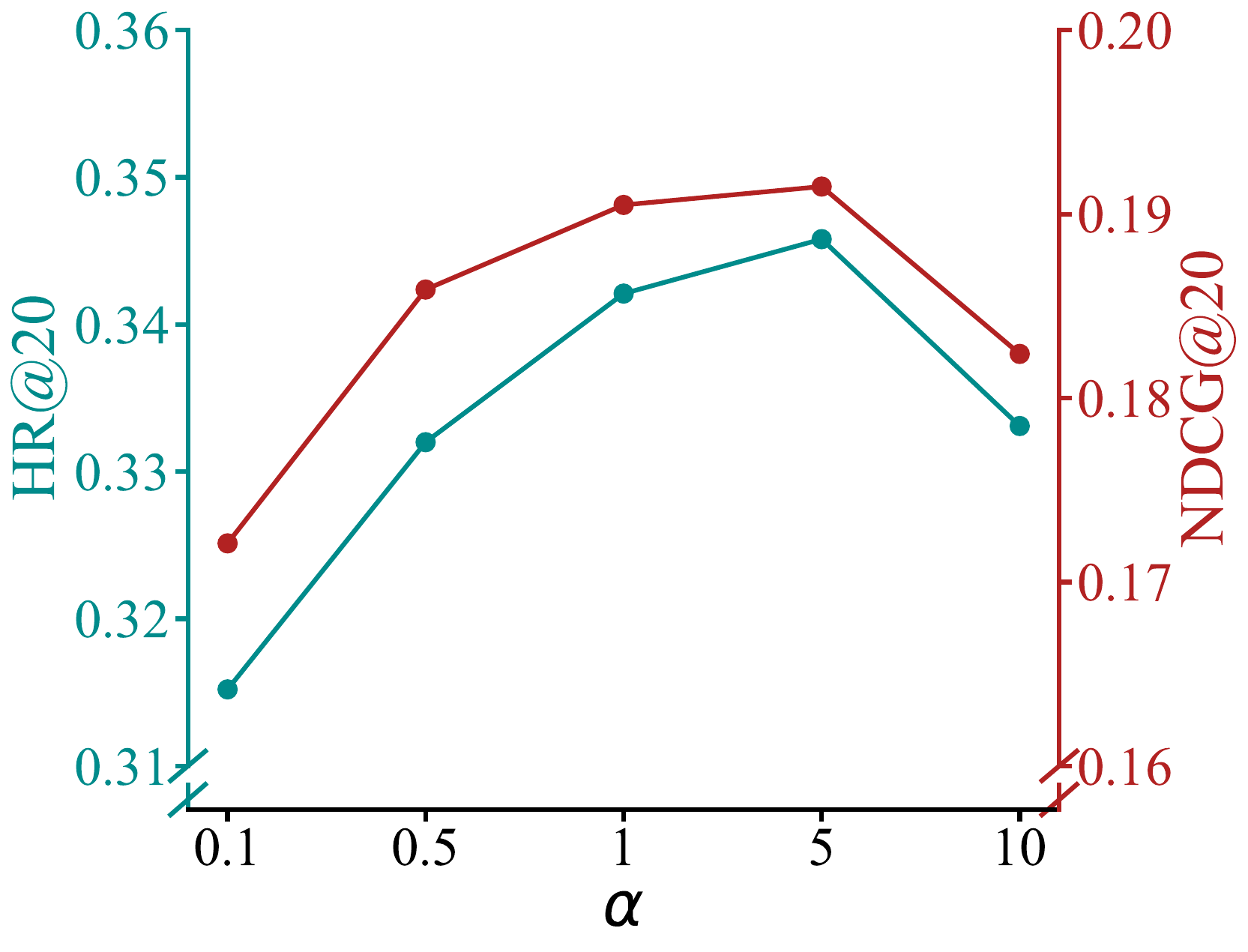}
		\subcaption{Tmall}
	\end{subfigure}
	\hfill
	\begin{subfigure}[ht]{0.32\textwidth}
		\includegraphics[width=1.\textwidth]{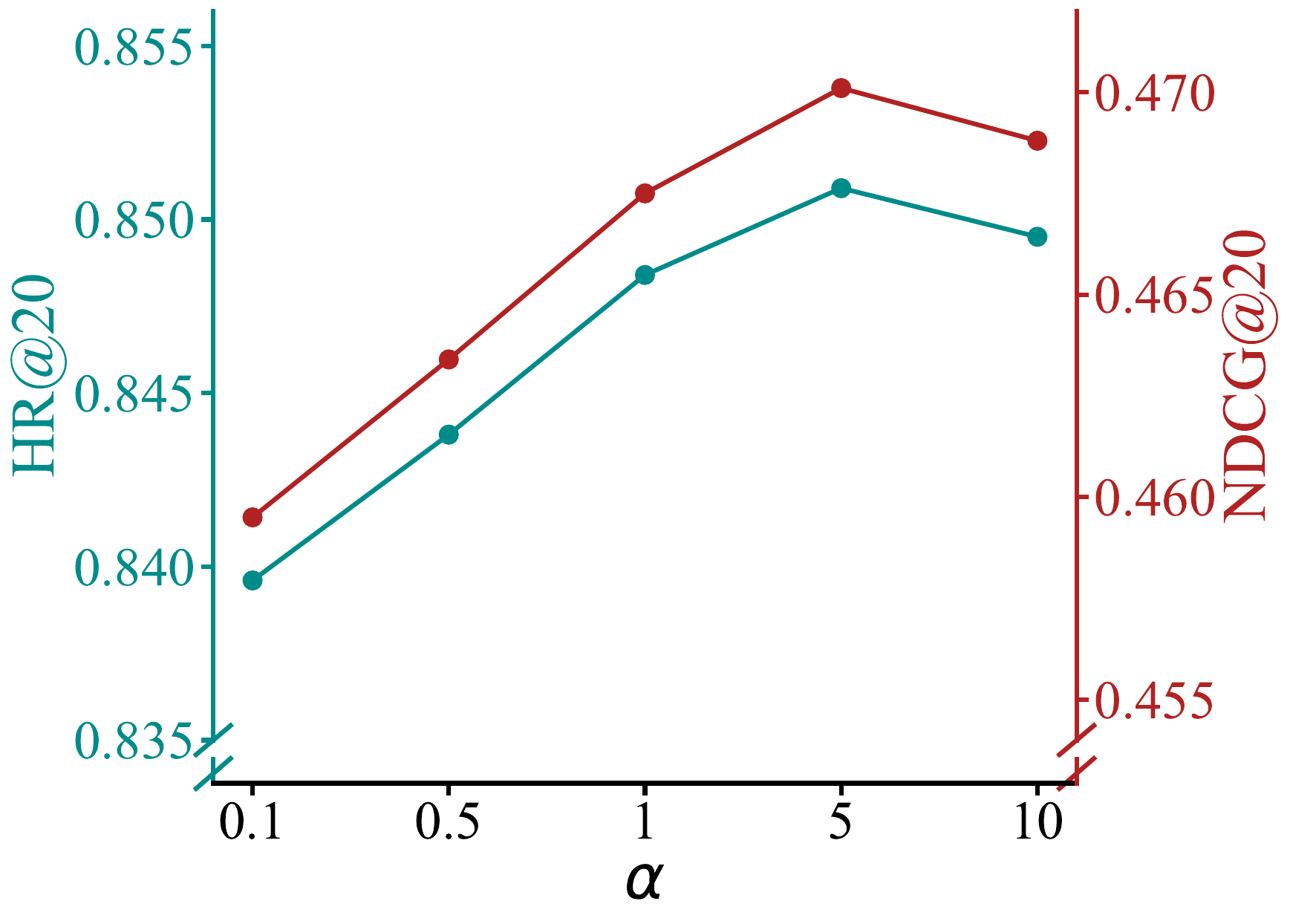}
		\subcaption{RL4RL-A}
	\end{subfigure}
	\caption{Effects of trade-off parameter $\alpha$ on three datasets with SR-GNN as the backbone.}
	\label{fig_param_alpha}
\end{figure}

\begin{figure}[ht]
	\centering
	\begin{subfigure}[ht]{0.32\textwidth}
		\includegraphics[width=1.\textwidth]{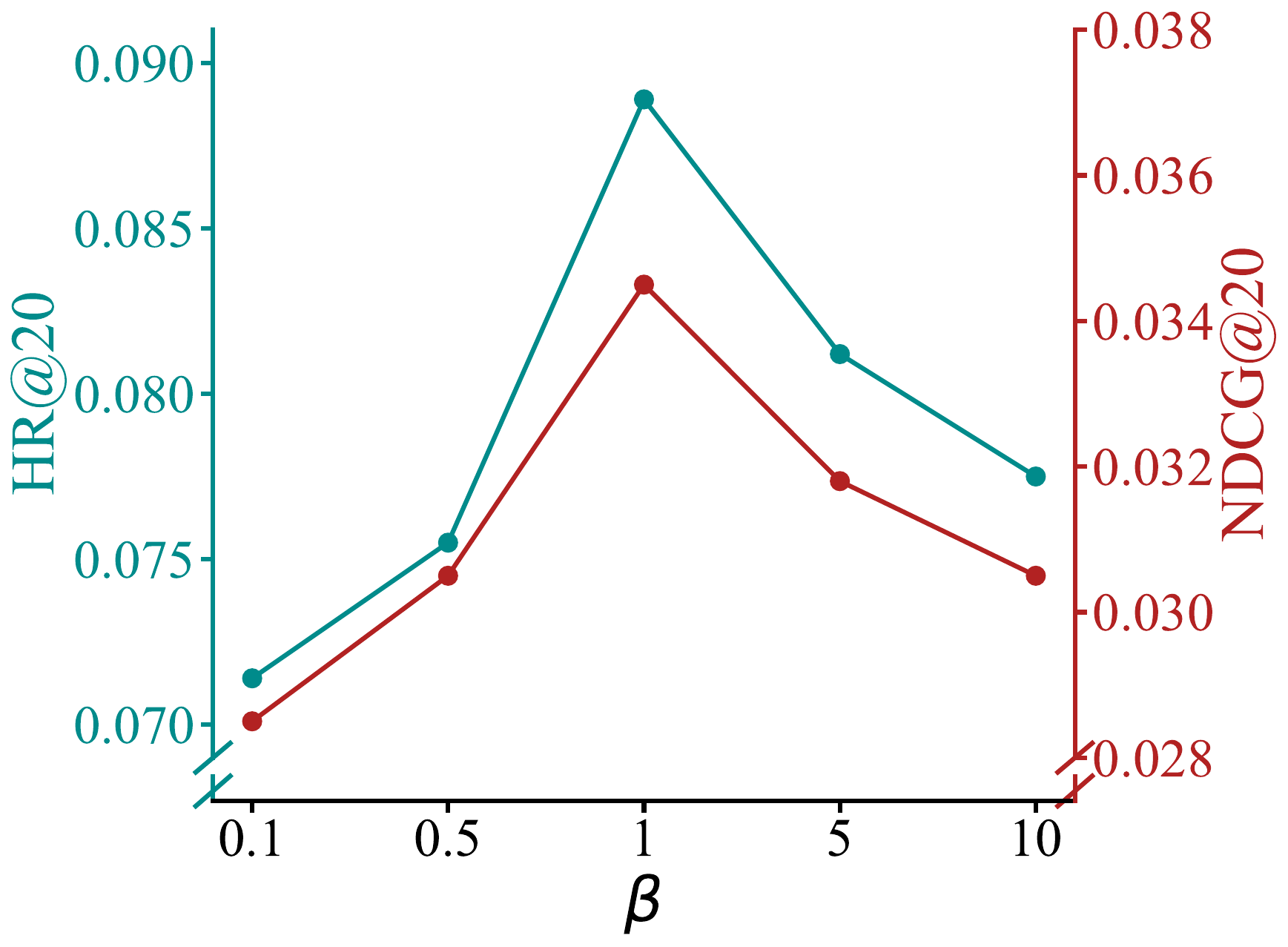}
		\subcaption{Yelp}
	\end{subfigure}
	\hfill
	\begin{subfigure}[ht]{0.32\textwidth}
		\includegraphics[width=1.\textwidth]{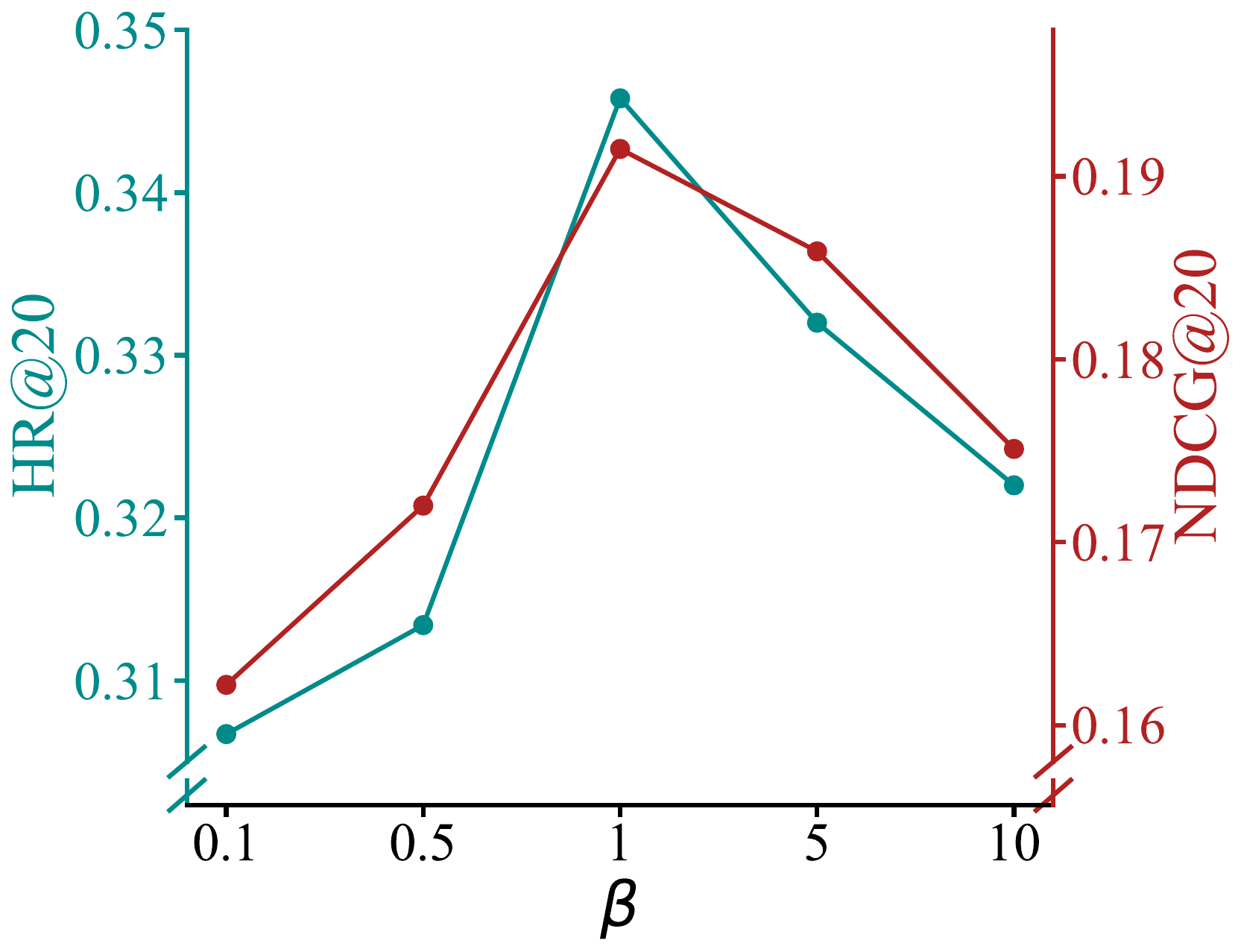}
		\subcaption{Tmall}
	\end{subfigure}
	\hfill
	\begin{subfigure}[ht]{0.32\textwidth}
		\includegraphics[width=1.\textwidth]{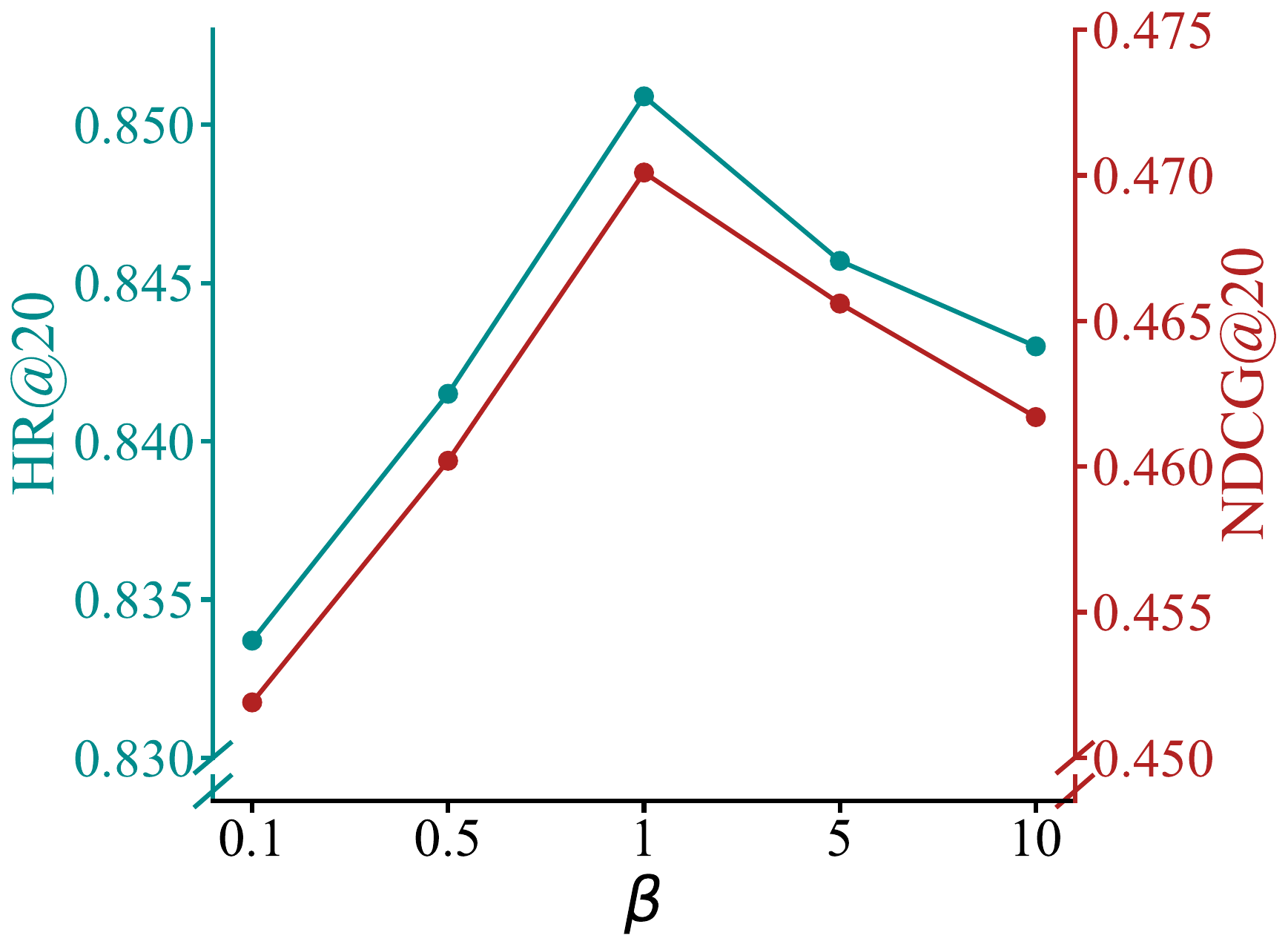}
		\subcaption{RL4RL-A}
	\end{subfigure}
	\caption{Effects of trade-off parameter $\beta$ on three datasets with SR-GNN as the backbone.}
	\label{fig_param_beta}
\end{figure}

	\subsubsection{Parameters Sensitivity\ \textbf{(RQ5)}}

\begin{figure}[ht]
	\centering
	\begin{subfigure}[ht]{0.45\columnwidth}
		\includegraphics[width=.9\textwidth]{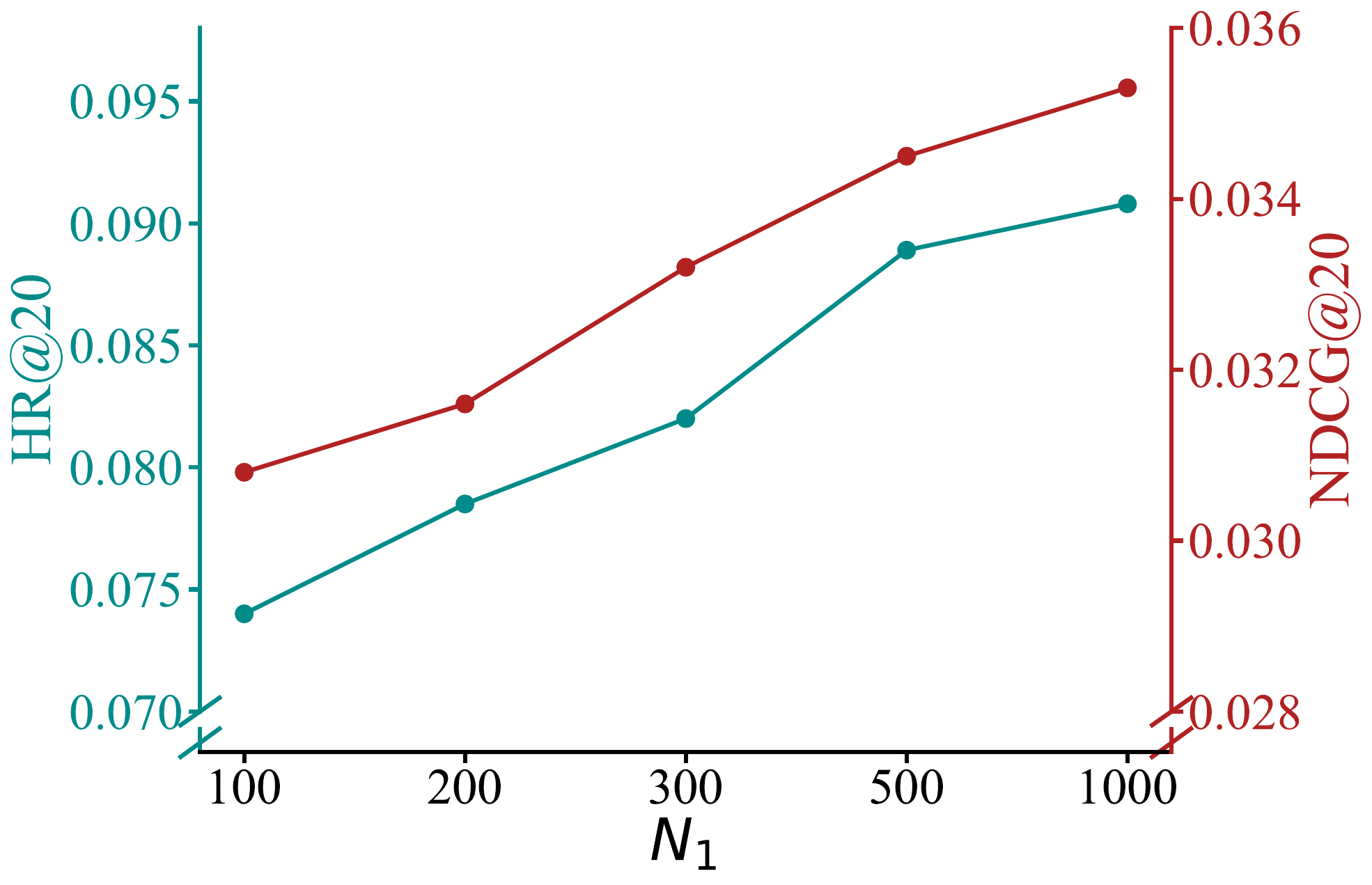}
		\subcaption{Yelp}
	\end{subfigure}
	\hfill
	\begin{subfigure}[ht]{0.45\columnwidth}
		\includegraphics[width=.9\textwidth]{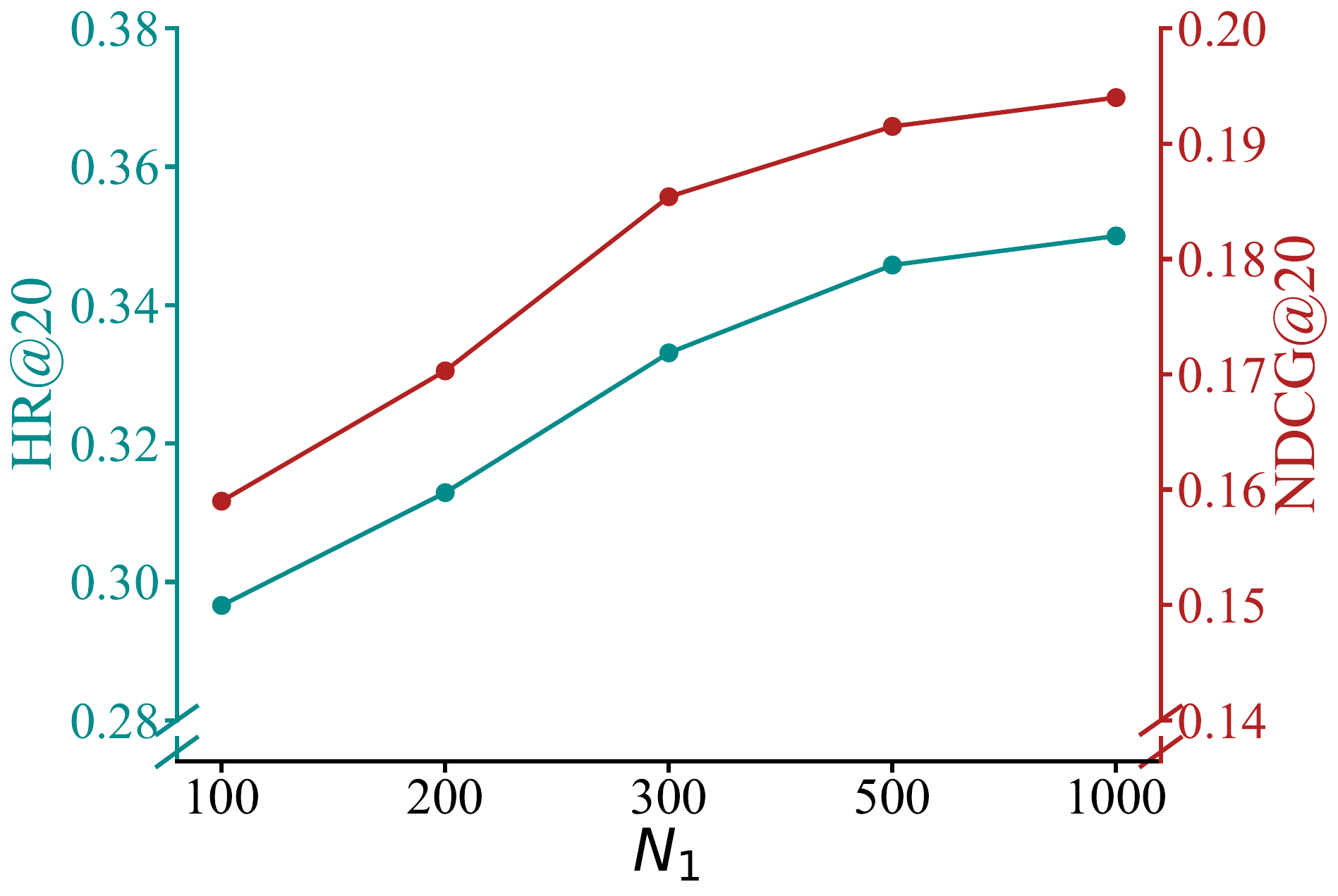}
		\subcaption{Tmall}
	\end{subfigure}
	\caption{Effect of $N_1$ in $\mathcal{L}_{REG}$ on Yelp and Tmall with SR-GNN as the backbone.}
	\label{fig_param_k}
\end{figure}

	In this subsection, we conduct a series of experiments to reflect the empirical influences of several trade-off parameters in ECoC. We mainly report the impact of $\alpha$ in $\mathcal{L}_{critic}$, $\beta$ in $\mathcal{L}_{actor}$ and the numbers $N_1$ in $\mathcal{L}_{REG}$. The results are demonstrated in Figure~\ref{fig_param_alpha}, Figure~\ref{fig_param_beta} and Figure~\ref{fig_param_k} respectively. Note that since RL4RS-A contains insufficient items (i.e., 283), we do not study the effect of $N_1$ on this dataset.
	
	First, for parameter $\alpha$ on three datasets, we observe from Figure~\ref{fig_param_alpha} that the performance of ECoC is relatively robust when it is in a certain range, i.e., $\alpha \in [1,5]$ on Yelp and Tmall. However, on RL4RS-A, a larger $\alpha$ is preferred since we might have to impose stronger restraint to help the learning of the Q function, which again is caused by the spars number of candidate items.
	
	However, the broken lines regarding $\beta$ in Figure~\ref{fig_param_beta} display more fluctuant tendencies as the value varies, especially when the value is larger than 1.
	On one hand, similar to the implication from the ablation study, too small values of $\beta$ would hinder the representation learning of items, thus affecting imitation performance. On the other hand, larger values also adversely influence the results as the directional guidance from the critic is suppressed.
    Therefore, we suggest a value of around 1 in order to obtain a better balance between $\mathcal{L}_{DC}$ and $\mathcal{L}_{BC}$.
	
	Last but not least, the balance of efficiency and effectiveness during negative sampling is also under consideration, especially with the increasing scale of candidate items in practice. Here, we particularly study the number $N_1$ in the conservatism regularization term $\mathcal{L}_{RGE}$ in the critic.
	The outcomes in Figure~\ref{fig_param_k} present a clear tendency that as $N_1$ increases, the HR@20 and NDCG@20 both rise up at least for those $N_1$ smaller than 1000. However, the promotion has been marginal when we enlarge $N_1$ from 500 to 1000. Consequently, in order to balance the computational cost, we finally set $N_1$ to 500 in the experiments.


%% file: section-appendix.tex
\section{Constrained Deterministic Policy Gradient}\label{app_derive}
    In this appendix, we provide a complete derivation of constrained deterministic policy gradient.
    Given a fixed dataset $\mathcal{H}$, the core idea is to prevent severe distribution shifts for either policy evaluation or policy improvement.
    
    Under the ideal circumstance, we can derive an empirical MDP $\widehat{\mathcal{M}}$, with the evaluation on a class of constrained policies $\pi(\cdot)$ under this MDP being accurate.
	Formally, $\widehat{\mathcal{M}}$ shares the same $\mathcal{S}, \mathcal{A}, \mathcal{R}$ and $\gamma$ with the original MDP $\mathcal{M}$ defined in Section~\ref{rl_formula}.  The exclusive difference comes from the state transition function $\widehat{\mathcal{P}}$, where the permitted transitions are restrained to those encountered in dataset $\mathcal{H}$ and further proportional to the co-occurrence frequencies, i.e.,
    \begin{equation*}
    \begin{aligned}
        \widehat{\mathcal{P}} \left(s, a, s' \right) &= \mathcal{P} \left(s, a, s' \right)\mathbb{I}\left\{ \left(s, a \right) \in \mathcal{H} \right\} \\
        &= N\left(s, a, s' \right) / \sum\nolimits_{\tilde{s}} N\left(s, a, \tilde{s} \right),
    \end{aligned}
    \end{equation*}
    where $\mathbb{I}$ is an indicator function and $N\left(s, a, s' \right)$ is the frequency when $\left(s, a, s' \right)$ is jointly observed in $\mathcal{H}$.
    Meanwhile, a proper stochastic behavioral policy $\hat{\pi}_b(\cdot \mid s)$ can be estimated as
    \begin{equation}\label{eq_pi_b}
        \hat{\pi}_b(a \mid s) = N\left(s, a \right) / \sum\nolimits_{\tilde{a}} N\left(s, \tilde{a} \right).
    \end{equation}
    
    Intuitively, by means of shrinking both transition probability and policy spaces, the unfamiliar states and unobserved actions at the next timestep are fantastically avoided.
    As a result, \citet{fujimoto2019off} proved that for deterministic policies, the evaluation in the policy space $\left\{ \pi \mid \left(s, \pi(a) \right) \in \mathcal{H}, \forall s \in \mathcal{H} \right\}$ is unbiased, compared with the results obtained under the complete MDP, i.e.,
	\begin{equation}\label{eq_bcq}
        Q^{\pi, \mathcal{M}}(s, a) = Q^{\pi, \widehat{\mathcal{M}} } (s, a).
	\end{equation}

    Note that such a theoretical result is applicable to either discrete or continuous action spaces.
    For policies on discrete action spaces, \citet{fujimoto2019off} further remarked that a constrained greedy strategy over the $Q$ values exactly develops the optimal constrained policy, i.e., $\pi^{\ast}=\operatorname*{argmax}_{a\ \text{s.t.} (s, a)\ \in \mathcal{H}} Q^{\pi}(s, a)$.
    However, the update manner of continuous policies is still not discussed due to the problematic greedy operation.
    Here, we first extend the constrained update to the optimal policy in the continuous case.
    Following the deterministic policy gradient (DPG) theorem~\cite{silver2014deterministic} for complete MDP $\mathcal{M}$, we make it hold for the empirical MDP $\widehat{\mathcal{M}}$.
    
    Roughly, to perform policy improvement, the chain gradient in terms of the continuous action function $Q(s, \pi(s))$ just indicates the direction that achieves the best gain when optimizing $\pi(s)$, even when such policy is constrained by the logged data.

    We start by recapping the core result of the DPG theorem. Formally, under some regularity conditions (see Condition A.1 in~\cite{silver2014deterministic}) which imply the existence of $\nabla _{\theta} \pi_{\theta}(s)$ and $\nabla_{a} Q^{\pi}(s, a)$, the gradient of the objective
    \begin{equation}
    	J(\pi_{\theta}, \mathcal{M})= \mathbb{E}_{s \sim \rho^{\pi}}[Q^{\pi}(s, \pi_{\theta}(s))]
    \end{equation}
    w.r.t. the parameters $\theta$ can be exactly represented as
    \begin{equation}\label{eq_dpg}
    	\nabla _{\theta} J(\pi_{\theta}, \mathcal{M}) = \mathbb{E}_{s \sim \rho^{\pi}} \left[\nabla _{\theta} \pi_{\theta}(s) \nabla_{a} Q^{\pi}(s, a) \mid _{a=\pi_{\theta}(s)} \right],
    \end{equation}
    where $\rho^{\pi}$ denotes the state distribution under policy $\pi$.

    For those policies constrained by offline data $\mathcal{H}$, after restraining the state distribution as $\hat{\rho}^{\pi}$, the performance objective can be rewritten as
    \begin{align}\label{eq_empirical_obj}
    	\begin{split}
    		J (\pi_{\theta}, \widehat{\mathcal{M}}) &=\mathbb{E}_{s \sim \rho^{\pi}} \left[Q^{\pi}(s, \pi_{\theta}(s)) \mathbb{I} \left\{(s, \pi_{\theta}(s)) \in \mathcal{H} \right\} \right] \\
    		&=\mathbb{E}_{s \sim \rho^{\pi}} \left[Q^{\pi}(s, \pi_{\theta}(s)) \mathbb{I} \left\{s \in \mathcal{H} \right\} \mathbb{I} \left\{\pi_{\theta}(s) \in \mathcal{H} \right\} \right] \\
    		&=\mathbb{E}_{s \sim \hat{\rho}^{\pi}} \left[Q^{\pi, \mathcal{H}}(s, \pi_{\theta}(s)) \right].
    	\end{split}
    \end{align}
    However, $Q^{\pi, \mathcal{H}}(s, a)$ might not be a globally differentiable function since we do not define the values for $(s,a) \notin \mathcal{H}$, thus the existence of the gradient at $a=\pi _{\theta}(s)$ is not guaranteed.
    To solve this, we moderately extend the definition of $Q^{\mathcal{H}}$ with $Q^{\mathcal{H}}_{\phi}$, where the latter is the practical function approximator (parameterized by $\phi$). Specifically, for those actions in the $\epsilon$-neighbourhood of $a$, we have
    \begin{equation}
    	Q^{\mathcal{H}}(s,a)=Q^{\mathcal{H}}_{\phi}(s,a), \quad \forall a \in \mathcal{A}_{a,\epsilon},
    \end{equation}
    where $\mathcal{A}_{a,\epsilon} =\left\{ a' \in \mathcal{A} : \| a' - a \| \leq \epsilon \right\}$.
    With the expanded $Q^{\mathcal{H}}(s,a)$, the chain rule regarding the gradient of $J (\pi,  \widehat{\mathcal{M}})$ still holds, which means we have
\begin{align}\label{eq_gradient_derive}
	\begin{split}
		\nabla _{\theta} J (\pi_{\theta},  \widehat{\mathcal{M}}) &= \mathbb{E}_{s \sim \rho^{\pi}} \left[\nabla _{\theta} Q^{\pi}(s, \pi_{\theta}(s)) \mathbb{I} \{(s, \pi_{\theta}(s)) \in \mathcal{H} \} \right] \\
		&\ (\text{Applying} \ \text{Eq.}~\eqref{eq_dpg}) \\
		&=\mathbb{E}_{s \sim \rho^{\pi}}  \left[ \nabla _{\theta} \pi_{\theta}(s) \nabla_{a} Q^{\pi}(s, a)|_{a=\pi_{\theta}(s)}  \mathbb{I} \left\{(s, \pi_{\theta}(s)) \in \mathcal{H} \right\} \right] \\
        &=\mathbb{E}_{s \sim \rho^{\pi}} \left[\nabla _{\theta} \pi_{\theta}(s) \mathbb{I} \{\pi_{\theta}(s) \in \mathcal{H} \} \nabla_a Q^{\pi}(s,a) | _{a=\pi_{\theta}(s) \mathbb{I} \{\pi_{\theta}(s) \in \mathcal{H} \}} \right] \\
		&= \mathbb{E}_{s \sim \hat{\rho}^{\pi}} \left[\nabla _{\theta} \pi ^{\mathcal{H}}_{\theta}(s) \nabla_{a} Q^{\pi, \mathcal{H}}(s,a) | _{a=\pi _{\theta}(s)} \right].
	\end{split}
\end{align}
    So far, we have derived the gradient of policy improvement in terms of the vanilla reward signals.
    However, since we introduce the conservatism regularization term in Section~\ref{sec_critic_conserv}, the influence of $\mathcal{L}_{REG}$ is still required to be analyzed.

    Based on the Theorem 3.5 in \cite{kumar2020conservative}, we notice that $\mathcal{L}_{REG}$ in Eq.\eqref{eq_reg_vanilla} is exactly the deterministic case of 
    \begin{equation}
    	\mathbb{E}_{a \sim \pi(a \mid s)} Q (s, a) - \mathbb{E}_{a \sim \hat{\pi}_b (a \mid s)} Q (s, a),
	\end{equation}
    where $\hat{\pi}_b(\cdot \mid s)$ is the estimated behavioral policy in Eq.~\eqref{eq_pi_b}. Therefore, as a limiting result for this theorem, we claim that $\mathcal{L}_{REG}$ actually makes the Q function solved by the recursive Bellman iteration with an altered reward $r(s, a) - \alpha \left[ \frac{1}{\hat{\pi}_b (a \mid s)} \mathbb{I}(\pi (s) = a) - 1 \right]$. Furthermore, the original performance objective is modified to be
    \begin{equation}\label{eq_reg_obj}
        J (\pi_{\theta}, \widehat{\mathcal{M}}) = \mathbb{E}_{s \sim \hat{\rho}^{\pi}} \left[Q^{\pi, \mathcal{H}}(s, \pi_{\theta}(s)) \right] - \alpha \frac{1}{1-\gamma} \mathbb{E}_{s \sim \hat{\rho}^{\pi}} \left[ \frac{1}{\hat{\pi}_b (\pi_{\theta}(s) \mid s )} - 1 \right].
    \end{equation}
    Compared with Eq.~\eqref{eq_empirical_obj}, the additional term in Eq.\eqref{eq_reg_obj} does not contribute to the final gradient in $\nabla _{\theta} J (\pi_{\theta},  \widehat{\mathcal{M}})$, leaving the gradient remains the form described in Eq.~\eqref{eq_gradient_derive}. Now we can conclude to prove the corollary~\ref{corollary_5.1}.